\pdfoutput=1
\documentclass[10pt,twocolumn,letterpaper]{article} 

\usepackage{cvpr}

\usepackage{times}
\usepackage{relsize}
\usepackage[T1]{fontenc} 
\usepackage[latin1]{inputenc} 
\usepackage[english]{babel}

\usepackage{epsfig}
\usepackage{graphicx}
\usepackage{wrapfig}
\usepackage[belowskip=0pt,aboveskip=0pt,font=small]{caption}
\usepackage[belowskip=0pt,aboveskip=0pt,font=small]{subcaption}
\usepackage{amsmath, amsthm, amssymb}
\usepackage{textcomp}
\usepackage{stmaryrd}
\usepackage{upgreek}
\usepackage{bm}
\usepackage{cases}
\usepackage{mathtools}

\usepackage{cite}
\usepackage[pagebackref=true,breaklinks=true,letterpaper=true,colorlinks,bookmarks=false,citecolor=red]{hyperref}

\usepackage{algorithm2e}

\usepackage{multirow}
\usepackage{rotating}
\usepackage{booktabs}

\usepackage{enumitem}
\usepackage[olditem,oldenum]{paralist}

\usepackage{alltt}
\usepackage{listings}
\belowdisplayskip \abovedisplayskip

\newlength{\sectionReduceTop}
\newlength{\sectionReduceBot}
\newlength{\subsectionReduceTop}
\newlength{\subsectionReduceBot}
\newlength{\abstractReduceTop}
\newlength{\abstractReduceBot}
\newlength{\captionReduceTop}
\newlength{\captionReduceBot}
\newlength{\subsubsectionReduceTop}
\newlength{\subsubsectionReduceBot}

\newlength{\eqnReduceTop}
\newlength{\eqnReduceBot}

\newlength{\horSkip}
\newlength{\verSkip}

\newlength{\figureHeight}
\usepackage{url}
\usepackage{xspace}
\usepackage{comment}
\usepackage{color}
\usepackage{afterpage}
\usepackage{pdfpages}
\usepackage{framed}
\usepackage{fancybox}
\usepackage[normalem]{ulem}
\usepackage{dsfont}

\usepackage{mysymbols}

\usepackage{fancyvrb}
\usepackage{booktabs}
\usepackage{paralist}
\usepackage{wrapfig}
\usepackage{rotating}

\newcommand{\djc}[1]{\sethlcolor{yellow}  \hl{\textbf{\texttt{Djc: #1}}}}

\definecolor{lightblue}{rgb}{.60,.75,1}

\newcommand{\xhdr}[1]{{\vspace{2pt}\noindent\textbf{\textit{#1}}}}
\newcommand{\rpm}{\raisebox{.2ex}{$\scriptstyle\pm$}}
\newcommand{\spm}[1]{\scriptsize{$\rpm$#1}}

\definecolor{darkorchid}{rgb}{0.5, 0, 0.5}

\newcommand{\aspace}{\hspace{10pt}}
\newcommand{\sspace}{\hspace{32pt}}
 \cvprfinalcopy 

\def\cvprPaperID{650} 

\ifcvprfinal\fi

\begin{document}

\title{
Why $M$ Heads are Better than One: \\Training a Diverse Ensemble of Deep Networks}
\author{
Stefan Lee$^1$ \aspace%
Senthil Purushwalkam$^2$ \aspace%
Michael Cogswell$^3$ \aspace%
David Crandall$^1$ \aspace%
Dhruv Batra$^3$\\
Indiana University$^1$ \sspace Carnegie Mellon University$^2$ \sspace Virginia Tech$^3$\\
{\tt\small \{steflee, djcran\}@indiana.edu \aspace spurushw@andrew.cmu.edu \aspace \{cogswell, dbatra\}@vt.edu}
}

\maketitle


\begin{abstract} 
Convolutional Neural Networks have achieved state-of-the-art
performance on a wide range of tasks. Most benchmarks
are led by ensembles of these powerful learners, but ensembling
is typically treated as a post-hoc procedure implemented by averaging
independently trained models with model variation induced by bagging
or random initialization. In this paper, we rigorously treat ensembling
as a first-class problem to explicitly address the question: what are the best
strategies to create an ensemble? We first compare
a large number of ensembling strategies, and then propose and
evaluate novel strategies, such as parameter sharing (through a new family of models
we call TreeNets) as well as training under ensemble-aware and
diversity-encouraging losses.  We demonstrate that TreeNets can 
improve ensemble performance and that diverse ensembles
can be trained end-to-end under a unified loss, achieving significantly higher ``oracle'' accuracies than classical ensembles.
\end{abstract} 


\section{Introduction}
\label{sec:intro}

Convolutional Neural Networks (CNNs) have shown impressive performance
on a wide range of computer vision tasks. An important (and perhaps
under-acknowledged) fact is that the state-of-the-art models are
generally \emph{ensembles} of CNNs, including nearly all of the top performers
on the ImageNet Large Scale Visual Recognition
Challenge~\cite{russakovsky2014imagenet}.  For example,
GoogLeNet~\cite{szegedy2014going}, one of the best-performing models
submitted to the ILSVRC challenge, is an ensemble achieving a five
percentage point increase in accuracy over a single base model of the
same architecture.

In these ensembles, multiple classifiers are trained to perform the
same task and their predictions are averaged to generate a new,
typically more accurate, prediction.  A number of related justifications
have been given for the success of ensembles, including:

\begin{compactenum}[i)]
\item Bayesian Model Averaging, that ensembles 
are a finite sample approximation to integration over the model class~\cite{domingos_icml00,minka_tr02,monteith_ijnn11};

\item Model Combination,  that ensembles enrich the space of hypotheses considered by the 
base model class and are representationally richer \cite{domingos_kdd97}; and

\item Reducing Estimation and Optimization Errors,
that ensemble averaging reduces the variance of base models, averaging
out variations due to objective function non-convexity, 
initialization, and stochastic learning~\cite{dietterich_mcs00,polikar2012ensemble}. 
\end{compactenum}

At the heart of these arguments is the idea of \emph{diversity}:
if we train multiple learners with decorrelated
errors, their predictions can be averaged to improve
performance~\cite{bishop_book}. In this work, we rigorously treat
ensembling as a problem in its own right, examining
multiple ensembling strategies ranging from standard bagging to
parameter sharing and ensemble-aware losses. We compare these methods
across multiple datasets and architectures, demonstrating that some
standard techniques may not be suitable for deep ensembles and
novel approaches improve performance.

\xhdr{Ensemble-Aware Losses.} Typically, ensemble members are trained
independently with no unifying loss, despite the fact that outputs are
combined at test time. It is common in classical literature to view
ensemble members as ``experts''~\cite{jacobs_nc91} or
``specialists''~\cite{hinton2014distilling}, but in typical practice
no effort is made to encourage diversity or specialization. It seems
natural then to question whether a ensemble-aware loss might result in
better performance. Here we study two ensemble-aware losses: (1)
directly training an ensemble to minimize the loss of the ensemble
mean, and (2) generalization of Multiple Choice
Learning~\cite{rivera_nips12} to explicitly encourage diversity.

\xhdr{Parameter Sharing.}  As a number of papers have demonstrated,
initial layers of CNNs tend to learn simple, generic features which vary
little between models, while deeper layers
learn features specific to a particular task and input
distribution
\cite{zeiler2014visualizing,girshick14CVPR,lenc2014understanding}. We
propose a family of tree-structured deep networks (which we call TreeNets) that
exploit the generality in lower layers by sharing them across
ensemble members to reduce parameters. We investigate the depth
at which sharing should happen, along a spectrum from single models
(full sharing) to independent ensembles (no sharing). This
coupling of lower layers naturally forces any diversity between
ensemble members to be concentrated in the deeper, unshared
layers. Perhaps somewhat surprisingly, we find that the optimal
setting is \emph{not} a classical ensemble, 
but instead a TreeNet that shares a few (typically 1-2) initial
layers. Thus tree-structured networks are a simple way to improve
performance while reducing parameters.

\xhdr{Model-Distributed Training of Coupled Ensembles.}
Unfortunately, both of the above approaches to coupling ensemble
members, either at the ``top'' of the architecture with ensemble-aware
losses that operate on outputs from all ensemble members, or at the
``bottom'' with parameter sharing in TreeNets, create significant
computational difficulties. Since networks are not independent, it is
no longer possible to train them separately in parallel, and sequential training may require
months of GPU time even for relatively small ensembles. Moreover,
even if training time is not a concern, larger models often take up
most of the available RAM on a GPU, so it is not possible to
fit an ensemble on one GPU. To overcome these hurdles, we present and will release a
novel MPI-based model-parallel distributed modification to the popular 
Caffe deep learning framework\cite{caffe} that implements cross-process communication 
as layers of the CNN.

We thoroughly evaluate each methodology across multiple datasets and
network architectures. These experiments 
cast new light on ensembles in deep networks, demonstrating
the effects of randomization in parameter and data space, parameter
sharing, and unified losses on modern scale vision problems. More
concretely, we:
\begin{compactenum}[i)]
\item rigorously treat CNN ensembling as its own problem,
\item introduce a family of models called TreeNets that permit a spectrum of degrees of layer-sharing,
\item present ensemble-aware and diversity-encouraging loses, and
\item present a distributed model-parallel framework to train deep ensembles.
\end{compactenum}

\section{Related Work}
\label{sec:related}

Neural networks, ensembles, and techniques to improve robustness and
diversity of grouped learners have decades of work in machine learning research,
but only recently have ensembles of
CNNs been studied. Related work can be broadly divided into two
categories: ensemble learning for general networks, and
its more recent application to CNNs.

\xhdr{Ensemble Learning Theory.} Neural networks have been applied in
a wide variety settings with many diverse modifications. Much of the
theoretical foundation for ensemble learning with neural networks was
laid in the 1990s. Krogh \textit{et al.}~\cite{krogh1995neural} and
Hansen and Salamon~\cite{hansen1990neural} provided theoretical and
empirical evidence that diversity in error distributions across member
models can boost ensemble performance.  This led to ensemble methods
that averaged predictions from models trained with different
initializations~\cite{hansen1990neural} and from models trained on
different bootstrapped training sets
\cite{zhou2002ensembling,krogh1995neural}. These methods take an
indirect approach to introducing diversity in ensembles.  Other work
has explicitly trained decorrelated ensembles of neural networks by
penalizing positive correlation between error
distributions~\cite{rosen1996ensemble,lee2012new,alhamdoosh2014fast}. While
effective on shallow networks, these methods have
not been applied to deeper architectures.

Although initially proposed for Structured SVMs, the work of
Guzman-Rivera et
al.~\cite{rivera_nips12,rivera_aistats14,rivera_relocal14} on 
Multiple Choice Learning (MCL) provides an attractive alternative
that does not require computing correlation between error. Related
ideas were studied by Dey et al.\cite{dey_rss12} in the context of
submodular list prediction. We generalize MCL and apply it to CNNs --
incorporate it with stochastic gradient descent-based training.

\xhdr{CNN Ensembles.} While ensembles of CNNs have been used
extensively, little work has focused on improving the ensembling
process. Most CNN ensembles 
use multiple random initializations or
training data subsets to inject diversity.
For example, popular ensembles of
VGG\cite{simonyan2014very} and AlexNet\cite{krizhevsky_nips12} simply
retrain with different initializations and average
the predictions. GoogLeNet\cite{szegedy2014going} induces diversity
with straightforward bagging, training each model with a
sampled dataset. Other networks, like Sequence to Sequence
RNNs~\cite{sutskever2014sequence}, use both approaches
simultaneously.

Parameter sharing is not a novel development in CNNs, but its
effect on ensembles has not been studied. Recent related work by
Bachman \textit{et al.}~\cite{bachman2014learning} proposed a general
framework called pseudo-ensembles for training robust models. They
define a pseudo-ensemble as a group of child models which are
instances of a parent model perturbed by some noise process. They
explicitly encourage correlation in model parameters through 
the parent  by a Pseudo-Ensemble Agreement (PEA) regularizer.
Although outwardly related to parameter sharing, pseudo-ensembles are
fundamentally different than the techniques presented here, as they use
parameter sharing to train a single robust CNN model rather than to
produce an ensemble with fewer parameters. Other recent work by Sercu
et al.~\cite{sercu2015very} uses parameter sharing in the context of
multi-task learning to build a common representation for multilingual
translation. Finally, Dropout~\cite{hinton2014distilling} can be interpreted as a procedure that trains an exponential number of highly related networks and cheaply combines them into one network, similar to PEA

One relevant recent work is~\cite{hinton2014distilling}, which briefly focuses
on ensembles. Members of this types of ensemble are specialists which
are trained on subsets of all possible labels with each subset manually designed
to include easily confused labels. These models are fine-tuned from one shared
generalist and then combined to make a final prediction. In contrast, our 
diversity-encouraging loss require no human hand-designing of class specialization 
-- our loss naturally allows members to specialize according to subset of classes 
or pockets of feature space, providing an end-to-end way of learning diverse ensembles.

\section{Experimental Design}
\label{sec:design}
We first describe the datasets, architectures, and evaluation
metrics that we use in our experiments to better understand
ensembling in deep networks.

\subsection{Datasets and Architectures}
We evaluate on three popular image classification benchmarks: CIFAR10
\cite{cifar10}, CIFAR100\cite{cifar10}, and the 2012 ImageNet Large
Scale Visual Recognition Challenge (ILSVRC) \cite{ilsvrc12}.
Since our goal is not to present new designs and architectures for
these tasks but rather to study the effect of different ensembling
techniques, unless otherwise noted we use standard models and training
routines.  All models are trained via stochastic gradient descent with
momentum and weight decay.

\xhdr{CIFAR10.} For this dataset, we use Caffe ``CIFAR10 Quick''~\cite{caffe} network 
as our base model. The reference model is trained using a batch 
size of 350 for 5,000 iterations with a momentum of 0.9, weight 
decay of 0.004, and an initial learning rate of 0.001 which drops 
to 0.0001 after 4000 iterations. We refer to this network and training 
procedure as CIFAR10-Quick.

\xhdr{CIFAR100.} We use the Network in Network model by Lin
et al.~\cite{LinCY13} as well as their reference training procedure,
which 
runs for 300,000 iterations with a batch size of 128,
momentum of 0.9, weight decay of 0.0001, and an initial learning rate
of 0.1. The learning rate decays by a factor of 10 whenever the
training loss fails to drop by at least 1\% over 20,000
iterations; this occurs twice over the course of a typical training run.
Our reference model's accuracy is about 4\% lower than
reported in~\cite{LinCY13}, because we do not perform their dataset
normalization procedure. We refer to this network and training procedure as
CIFAR100-NiN.

\xhdr{ILSVRC2012.} For this dataset we use both  the Network in Network
model~\cite{LinCY13} and CaffeNet (similar to AlexNet\cite{krizhevsky_nips12}). Both networks are
trained for 450,000 iterations with an initial learning rate of 0.1,
momentum of 0.9, and weight decay of 0.0005. For NiN the batch size is
128 and the learning rate is reduced by a factor of 10 every 200,000
iterations. For CaffeNet the batch size is 256 and the learning rate
schedule is accelerated, reducing every 100,000 iterations. We 
refer to these models as ILSVRC-NiN and ILSVRC-Alex, respectively.

\subsection{Evaluation Metrics}
We evaluate our ensemble performance with respect to two different metrics.
\textbf{Ensemble-Mean Accuracy} is the accuracy of
the ``standard'' test-time procedure for ensembles --
 averaging the
beliefs of all members and predicting the most confident class.
Strong performance on this metric indicates that the ensemble members
generally agree on the correct response, with errors reduced by
smoothing across members. In contrast, \textbf{Oracle Accuracy} is the
accuracy of the ensemble if an ``oracle'' selects the prediction of
the \emph{most accurate} ensemble member for each example. Oracle
Accuracy demonstrates what the ensemble knows as a collection of
specialists, and has been used in prior work to measure ensemble
performance~\cite{rivera_nips12,rivera_aistats14,rivera_relocal14,dey_rss12,batra2012diverse}.


\section{Random Initialization and Bagging}
\label{sec:stand}

We now present our analysis of different approaches to training CNN
ensembles. This section focuses on standard approaches, while
Sections \ref{sec:tree} and \ref{sec:aware} present novel ideas on
parameter sharing and ensemble-aware losses.

Randomly initializing network weights and randomly re-sampling dataset subsets
(bagging) are perhaps the most commonly-used methods to create model
variation in members of CNN ensembles. 
Table \ref{tab:bag} presents results
using three different ensembling techniques: (1) Random Initialization, in which
all member models see the same training data but are initialized using different
random seeds, (2)Bagging, in which each member uses the same initial
weights but trains on a subset of data sampled (with replacement) from the original,
and (3) Combined, which uses both techniques.
Numbers in the table are accuracies and standard deviations across three trials.
The CIFAR ensembles were built with four members
while the ILSVRC ensembles had five.

\begin{table}[b]
\centering
\resizebox{\columnwidth}{!}{
\begin{tabular}{l c c  c c c}\toprule
              & \multicolumn{2}{c}{\textbf{Single Model}} & \textbf{Random Init.}        & \textbf{Bagging}             & \textbf{Combined} \\
\midrule \vspace{-12pt}& & &\\
& \multicolumn{2}{c}{Accuracy} & \multicolumn{3}{c}{Ensemble-Mean Accuracy} \\\midrule
\textbf{CIFAR10-Quick} \footnotesize{$\times$4} & \multicolumn{2}{c}{77.06 \spm{0.27}} & 80.72 \spm{0.10}  & 78.40 \spm{0.28}  & 78.95 \spm{0.17}  \\
\textbf{CIFAR100-NiN} \footnotesize{$\times$4} & \multicolumn{2}{c}{60.19 \spm{0.49}} & 66.51 \spm{0.27}  & 62.11 \spm{0.24}  & 61.73 \spm{0.16}  \\
\textbf{ILSVRC-Alex}  \footnotesize{$\times$5} & \multicolumn{2}{c}{56.79 \spm{0.04}} & 59.94 \spm{0.36}  & 57.46 \spm{0.12}  & 57.39 \spm{0.14}  \\
\textbf{ILSVRC-NiN}   \footnotesize{$\times$5} & \multicolumn{2}{c}{58.90 \spm{0.13}} & 64.08 \spm{0.11}  & 55.02 \spm{0.15}  & 60.51 \spm{0.12}  \\
\midrule \vspace{-12pt}& & &\\
& \multicolumn{2}{c}{Accuracy} & \multicolumn{3}{c}{Oracle Accuracy} \\\midrule
\textbf{CIFAR10-Quick} \footnotesize{$\times$4} & \multicolumn{2}{c}{77.06 \spm{0.27}} & 89.89 \spm{0.17}  & 89.94 \spm{0.27}  & 89.28 \spm{0.25}  \\
\textbf{CIFAR100-NiN} \footnotesize{$\times$4} & \multicolumn{2}{c}{60.19 \spm{0.49}} & 78.63 \spm{0.31}  & 75.47 \spm{0.12}  & 75.21 \spm{0.28}  \\
\textbf{ILSVRC-Alex}  \footnotesize{$\times$5} & \multicolumn{2}{c}{56.79 \spm{0.04}} & 70.45 \spm{0.63}  & 69.58 \spm{0.17}  & 69.61 \spm{0.17}  \\
\textbf{ILSVRC-NiN}   \footnotesize{$\times$5} & \multicolumn{2}{c}{58.90 \spm{0.13}} & 73.60 \spm{0.07}  & 67.79 \spm{0.02}  & 72.92 \spm{0.00}  \\ \bottomrule
\end{tabular}
}
\vspace{5px}
\caption{Comparison of standard ensembling techniques. All ensembles outperform their base models, but bagging shows smaller gains resulting from reduced training data.}
\label{tab:bag}
\end{table}

As expected, all ensembles improve performance over their single base
model. Somewhat surprisingly, we find that bagging reduces
Ensemble-Mean Accuracy compared to random initialization alone, while
Oracle Accuracy remains nearly constant. This
result suggests that the bagged networks are poorly calibrated, such
that confident incorrect responses are negatively impacting
results. The individual member networks (not shown in table) also
perform worse than those trained on the original dataset.  We
attribute these results to the reduction in unique training exemplars
that bagging introduces. Given an initial dataset of $M$ examples from
which we draw $M$ points with replacement to make a bagged set $B$,
the probability of an example $X_i$ being sampled at least once is
$P(X_i \in B)=1 - \left(\frac{(M-1)}{M}\right)^M$. The expected
fraction of examples drawn at least once is thus
$1-\left(1-\frac{1}{M}\right)^M$, which is approximately
$1-1/e \approx 0.63$ for large $M$; i.e. bagging costs over a third of
our unique data points! Not only are we losing 37\% of our data, we
are also introducing that many duplicated data points. To examine
whether these duplicates affect performance, we reran the CIFAR10
experiments with a dataset of 31,500 unique examples (approximately
63\% of the original dataset) and found similar reductions in accuracy,
indicating that the loss of unique data is the primary negative effect of bagging.

 Note that for convex or shallow models, the loss of unique
 exemplars in bagging is typically acceptable as random parameter
 initialization is simply insufficient to produce diversity. To the
 best of our knowledge, this is the first finding to
 establish \textbf{that random initialization may not only be
 sufficient but \emph{preferred} over bagging for deep networks given
 their large parameter space and the necessity of large training data}.

\section{Parameter Sharing with TreeNets}
\label{sec:tree}

Ensembles and single models can be seen as two endpoints on a
spectrum of approaches: single models require a careful
allocation of parameters to perform well, while ensembles extract as
much performance as possible from multiple instances of a base
model. Ensemble approaches likely introduce wasteful duplication of
parameters in generic lower layers, increasing training time and model
size. The hierarchical nature of CNNs makes them well-suited to
alternative ensembling approaches where member models benefit from
shared information at the lower layers while retaining the advantages
of classical ensembling methods.  

\begin{figure}[t]
\centering
\resizebox{0.98\columnwidth}{!}{
\includegraphics[clip=true,trim=5px 5px 15px 10px]{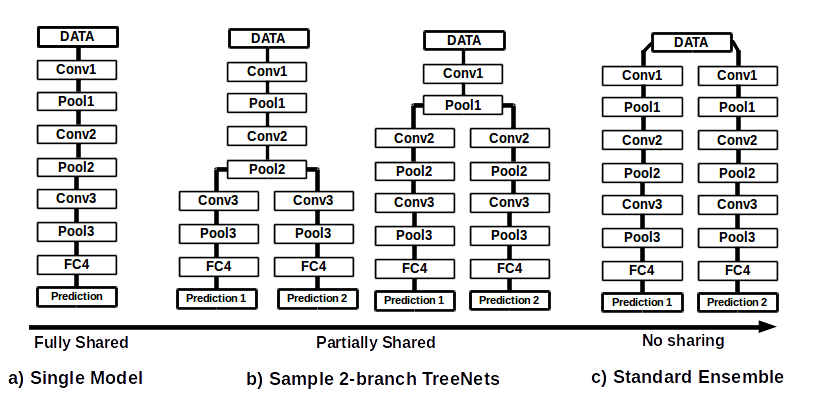}
}
\caption{TreeNets exist on a spectrum between single models and fully independent ensembles.}
\label{fig:tree}
\end{figure}

Motivated by this observation, in this section we present and evaluate a family of tree-structured
CNN ensembles called TreeNets, as shown in Figure~\ref{fig:tree}.
A TreeNet is an ensemble consisting of zero or more shared initial layers,
followed by a branching point and zero or more independent layers.
During training, the shared layers above a branch receive gradient
information from each child network, which are accumulated according
to back-propagation. At test time, each path from root to leaf can be
considered an independent network, except that redundant computations
at the shared layers need not be performed.

We evaluated our novel TreeNet models on the two larger architectures
trained on ImageNet, ILSVRC-Alex and ILSVRC-NiN, and Table~\ref{tab:tree}
presents the results. 
The table shows the Ensemble-Mean Accuracy (again in terms of means and standard
deviations across three trials) achieved by TreeNets with splits at different depths.
For example, splitting at conv2 means that all layers up to and including conv2
are shared, and all branches are independent afterwards.
Since layers that do not contain any
parameters (e.g. pooling, nonlinearity) are unaffected by parameter
sharing, we only show results for splitting on parameterized layers.

\begin{table}[t]
\centering
\resizebox{.52\columnwidth}{!}{
\begin{tabular}{c  c}\toprule
    \multirow{1}{*}{ILSVRC-Alex \footnotesize{$\times$5}} &   \textbf{Ensemble-Mean}     \\
    Split Point & \textbf{Accuracy}\\
\midrule \vspace{-12pt}& \\
{ensemble} & 59.47 \spm{0.45}\\
\textbf{conv1} & \textbf{59.62} \spm{0.09}\\
{conv2} & 59.32 \spm{0.17}\\
{conv3} & 58.39 \spm{0.10}\\
{conv4} & 57.73 \spm{0.05}\\
{conv5} & 55.25 \spm{0.03}\\
single model & 56.79 \spm{0.04}\\
 \bottomrule
\end{tabular}
}~
\resizebox{.465\columnwidth}{!}{
\begin{tabular}{c c}\toprule
    \multirow{1}{*}{ILSVRC-NiN \footnotesize{$\times$4}} &   \textbf{Ensemble-Mean}   \\
    Split Point& \textbf{Accuracy}\\
\midrule \vspace{-12pt}& \\
ensemble & 64.08 \spm{0.00} \\
{conv1} & 65.50  \spm{0.24}\\
\textbf{cccp1} & \textbf{65.69}  \spm{0.08}\\
{cccp2} & 65.64 \spm{0.11}\\
{conv2} & 65.64 \spm{0.07}\\
{cccp3} & 65.47 \spm{0.07} \\
{cccp4} & 65.62 \spm{0.01}\\
single model & 58.90  \spm{0.13}\\
 \bottomrule
\end{tabular}
}
\vspace{5px}
\caption{Results for TreeNet training at various depths of ILSVRC-Alex and ILSVRC-NiN. Ensemble performance is retained even with substantial parameter sharing.}
\label{tab:tree}
\end{table}

We see that \textbf{shared parameter networks not
only retain the performance of full ensembles, but can outperform
them}. For our best ILSVRC-NiN TreeNet, we improve accuracy
over standard ensembles while reducing the parameter count by 7\%.
It may be that lower layer representations, though
simple and generic, still had room for improvement. By sharing low
level weights, each weight is updated by multiple sources of
supervision, one per branch. This indicates TreeNets could provide
regularization which favors slightly better low level representations.

We find further evidence for this claim by looking at individual
branches of the TreeNet compared to the independently trained networks
of the ensemble. Regardless of split point, each TreeNet branch in our
shared ensemble achieved around 2 to 3 percentage points higher
accuracy than independent ensemble members. Unlike in classical
ensembles where each member model performs about as well as the base
architecture, TreeNets seem to boost performance of not only the
ensemble but the individual networks as well. We also experimented
with multiple splits leading to more complicated ``balanced binary''
tree structures on ILSVRC-NiN and found similar improvements.

We also tested ILSVRC-Alex TreeNet models trained for object detection
on PASCAL VOC 2007 \cite{pascal-voc-2007} dataset. We used the Fast
R-CNN~\cite{girshick14CVPR} architecture fine-tuned from our TreeNet
models. For the test-time bounding-box regression, we average the
results from each member model for an ensemble. We found a
statistically significant increase in mean average precision of about
0.7\% across multiple runs compared to starting from a standard
ensemble. We take these initial experiments to imply TreeNet models
are at least as generalizable to other tasks as standard
ensembles. More details are provided in the supplementary materials.

To summarize the key results in this section, we found that TreeNets
with a few (typically 1-2) initial layers outperform classical
ensembles, while also having fewer parameters which may reduce
test-time computation time and memory requirements.

\section{Training Under Ensemble-Aware Losses}
\label{sec:aware}
In the two previous sections, each ensemble member was trained with
the same objective -- independent cross-entropy of each ensemble
member. What happens if the objective is aware of the ensemble? We begin by showing a surprising result:
the first ``natural'' idea of simply optimizing the performance of the
average-beliefs of the ensemble does \emph{not} work, and we provide
intuitions why this is the case (lack of diversity).  This negative
result shows that a more careful design for ensemble-aware loss
functions is crucial. We then propose a diversity-encouraging loss function
that shows significantly improved oracle performance.

\subsection{Directly Optimizing for Model Averaging}
For a standard ensemble, test-time classification is typically
performed by averaging the output of the member networks, so it is
natural to explicitly optimize the performance of the corresponding
Ensemble-Mean loss during training. We ran all four ensemble
architectures under two settings: (1) Score-Averaged, in which we
average the last layer outputs (i.e.\ the scores that
are \emph{inputs} to the softmax function), and (2)
Probability-Averaged, in which we average the softmax probabilities
of ensemble members.  Intuitively, the difference between the two
settings is that the former assumes the
ensemble members are ``calibrated'' to produce scores of similar
relative magnitudes while the latter does not.

\begin{table}[t]
\centering
\resizebox{\columnwidth}{!}{
\begin{tabular}{c c c c}\toprule
              & \textbf{Independent Losses}        & \textbf{Score-Averaged}             & \textbf{Prob-Averaged} \\
\midrule \vspace{-12pt}& & &\\
& Ensemble-Mean&\multicolumn{2}{c}{Accuracy} \\\midrule
\textbf{CIFAR10-Quick} \footnotesize{$\times$4} & 80.72 \spm{0.10} & 79.32 \spm{0.02} & 77.10 \spm{0.16} \\
\textbf{CIFAR100-NiN} \footnotesize{$\times$4} & 66.51 \spm{0.27} & 65.77 \spm{0.21} & 62.77 \spm{0.28} \\
\textbf{ILSVRC-Alex}  \footnotesize{$\times$5} & 59.94 \spm{0.13} & 56.56 \spm{0.10} & 49.81 \spm{0.18} \\
\textbf{ILSVRC-NiN}   \footnotesize{$\times$5} & 83.43 \spm{0.10} & 79.24 \spm{0.36} & 42.39 \spm{0.24}\\
\bottomrule
\end{tabular}
}
\vspace{5px}
\caption{Results of training ensembles to reduce loss over member
  predictions averaged either over scores or probabilities.}
\label{tab:avg}
\vspace{-10pt}
\end{table}

Table \ref{tab:avg} shows the results of these experiments, again
averaged over three trials.  In all cases, network
averaging \textit{reduced} performance, with Probability-Averaged
causing greater degradation. This is counter-intuitive: explicitly
optimizing for the performance of Ensemble-Mean does \emph{worse} than
averaging independently trained models. We attribute this to two 
problems, which we now discuss: lack of diversity and
numerical instability.

\xhdr{Averaging Outputs Reduces Diversity.} Unfortunately, averaging scores or probabilities 
during training has the unintended consequence of eliminating
diversity in gradients back-propagated through the ensemble. Consider
a generic averaging layer, \vspace{-10pt}
$$ \mu(\boldsymbol{x_1},...,\boldsymbol{x_N})
= \frac{1}{N}\sum_{i=1}^N \boldsymbol{x_i},$$ that ultimately
contributes to some loss $\ell$, and consider the derivative of $\ell$
with respect to some $\boldsymbol{x_i},$
\[ \frac{\del\ell}{\del \boldsymbol{x_i}} =  \frac{\del\ell}{\del\mu}\frac{\del\mu}{\del \boldsymbol{x_i}} = \frac{\del\ell}{\del\mu} \frac{1}{N}.\]
This expression does not depend on $i$ --- gradients back-propagated
into all ensemble members are identical!  Due to the averaging layer,
responsibility for mistakes is shared, which eliminates gradient diversity.
This is different from the behavior of an ensemble of independently
trained networks, where each member receives a different gradient
depending on individual performance.  
(The averaging also scales the gradients, so in our 
experiments we compensate by increasing the learning rate by a factor of $N$; otherwise,
 we found learning tended to arrive at even worse solutions.)

\xhdr{Averaging Probabilities Is Unstable.} We attribute the further loss of
accuracy when averaging probabilities (versus scores) 
to increased numerical instability. The softmax function's derivative
with respect to its input is unstable for outputs near 0 or 1.
However, when paired with a cross-entropy loss, the derivative of the
loss with respect to softmax input reduces to a simple subtraction.
Unfortunately, there is no similar simplification for 
cross-entropy over an average of softmax outputs (see supplemental
materials for details). Optimization under these conditions
is difficult, causing loss at convergence for
Probability-Averaged networks to be nearly twice that of
Score-Averaged networks, and about the same as a single network.

Motivated by the finding that decreased diversity from optimizing
Ensemble-Mean leads to reduced performance, we next present an explicit
diversity-encouraging loss.

\subsection{Adding Diversity via Multiple Choice Learning}
We have so far discussed the role of ensemble diversity in the context
of model averaging; however, in many settings, generating multiple
plausible hypotheses may be preferred to producing a single
answer. Ensembles fit naturally into this space as they produce
multiple answers by design. However, independently trained models
typically converge to similar solutions, prompting the need to
optimize for diversity directly. In this section, we develop and
experiment with diversity encouraging losses and demonstrate their
effectiveness at specializing ensembles.

We build on \emph{Multiple Choice Learning}
(MCL)~\cite{rivera_nips12}, which we briefly recap here. Consider a
set of predictors $\{\theta_1, ..., \theta_M\}$ such that $\theta_m: x
\rightarrow P$ where $P$ is a probability distribution over some set
of labels, and a dataset $D$=$\{(x_1,y_1), ..., (x_N,y_N)\}$, where each
feature vector $x_i$ has a ground truth label $y_i$.  From the point
of view of an oracle that only listens to the most correct $\theta_m$,
the loss for an example $(x,y)$ is
$$\mathcal{L}_{set}(x,y) = \min_{m \in [1,M]} \ell\left(\theta_m(x), y\right),$$
which we will call the oracle set-loss. Intuitively, given that the
oracle will select the most correct predictor, the loss on any example
is the minimum loss over predictors. Alternatively, the oracle loss
can be interpreted as allowing a system to guess $M$ times, scoring an
example as correct if any guess is correct. Thus an
ensemble of $M$ predictors is directly comparable to the commonly used
top-$M$ metric used in many benchmarks (e.g.\ top-5 in
ILSVRC~\cite{ilsvrc12}).

We adapt this framework to the cross-entropy loss used for
training deep classification networks. Given a single predictor
$\theta_m$, the cross-entropy loss for example $(x,y)$ is
$$ \ell(x,y) = -\log\left(p_y^{\theta_m}\right),$$
where $p_y^{\theta_m}$ is the predicted probability of class $y$. Let
$\alpha_{mi}$ be a binary variable indicating whether predictor
$\theta_m$ has the lowest loss on example $(x_i,y_i)$. We can then
define a cross-entropy oracle set-loss over a dataset $D$, \vspace{-4pt}
$$\mathcal{L}_{set}(D) = \frac{1}{|D|} \sum_{(x_i, y_i) \in D}\sum_{m=1}^M -\alpha_{mi} \log{\left(p_{y_i}^{\theta_m}\right)}.$$
Notice that just like cross-entropy is an upper-bound on training
error, this expression is an upper-bound on the oracle training
error~\cite{rivera_nips12}. Guzman-Rivera et al.~\cite{rivera_nips12} presented a
coordinate descent algorithm for optimizing such an objective. Their
approach alternates between two stages: first, each data point is
assigned to its most accurate predictors, and then models are trained
until convergence using only the assigned examples. 

Even if done in parallel, training multiple CNNs to convergence for
each iteration is intractable. We thus interleave the
assignment step with batch updates in stochastic gradient descent. For
each batch, we pass the examples through the network, producing
probability distributions over the label space from each ensemble
member. During the backward pass, the gradient of the loss for each
example is computed with respect only to the predictor with the lowest
error on that example (with ties broken randomly). Pseudo-code is available in the
supplement.

So far we have assumed
that the oracle can select only one answer, i.e. $\sum_m \alpha_{mi} =
1$, however this can easily be generalized to select the
$k$ predictors with lowest loss such that $\sum_m \alpha_{mi} =
k$. Varying $k$ from one to the number of predictors trades
off between diversity and the number of training examples
each predictor sees, which affects both generalization and convergence.

\begin{figure*}[t]
\begin{subfigure}{0.5\columnwidth}
\centering
\includegraphics[clip=true, trim=0px 10px 35px 20px, scale=0.29]{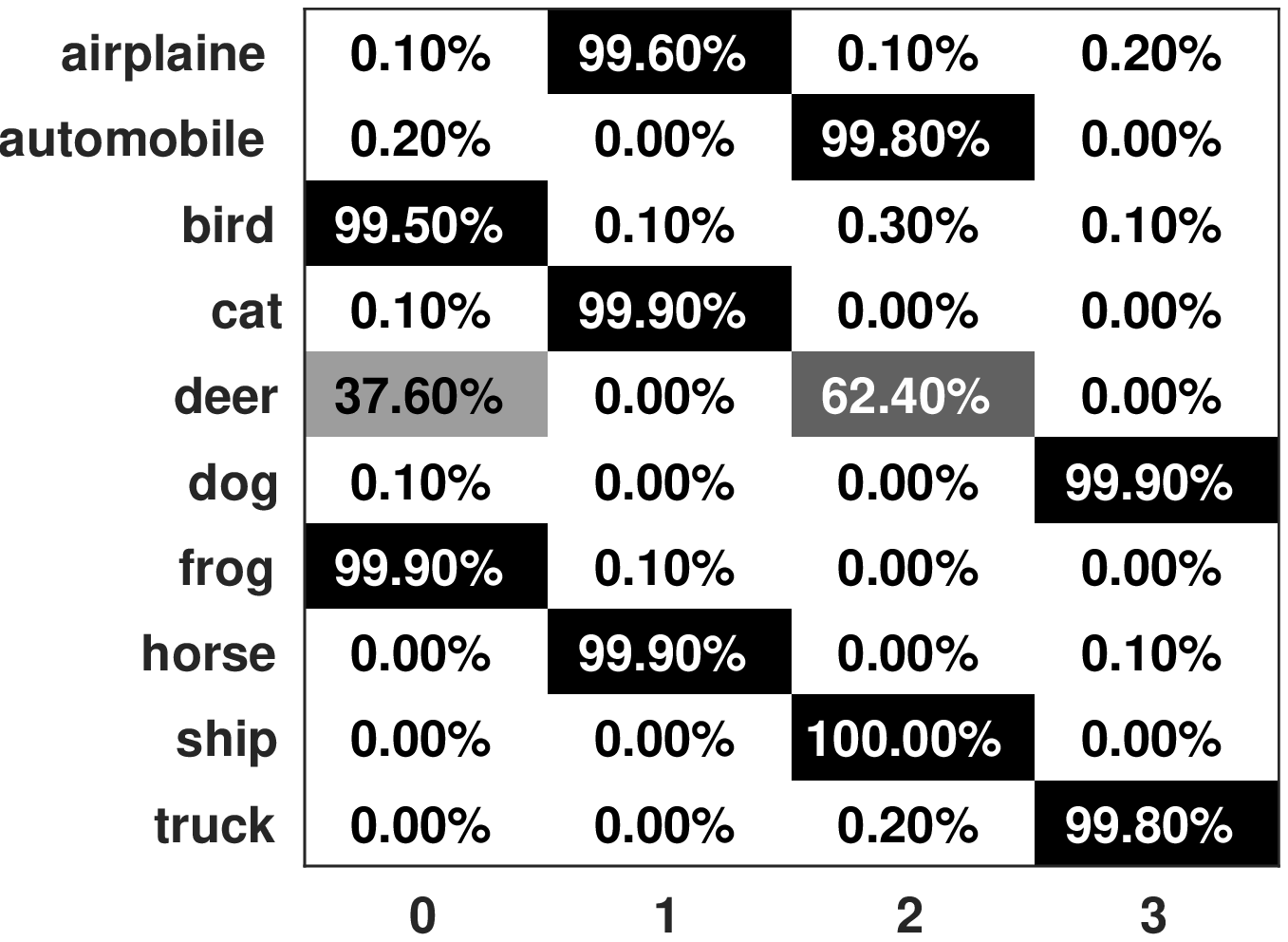}\null\hfill
\caption{k=1}
\end{subfigure}
\begin{subfigure}{0.5\columnwidth}
\centering
\includegraphics[clip=true, trim=40px 10px 35px 20px, scale=0.29]{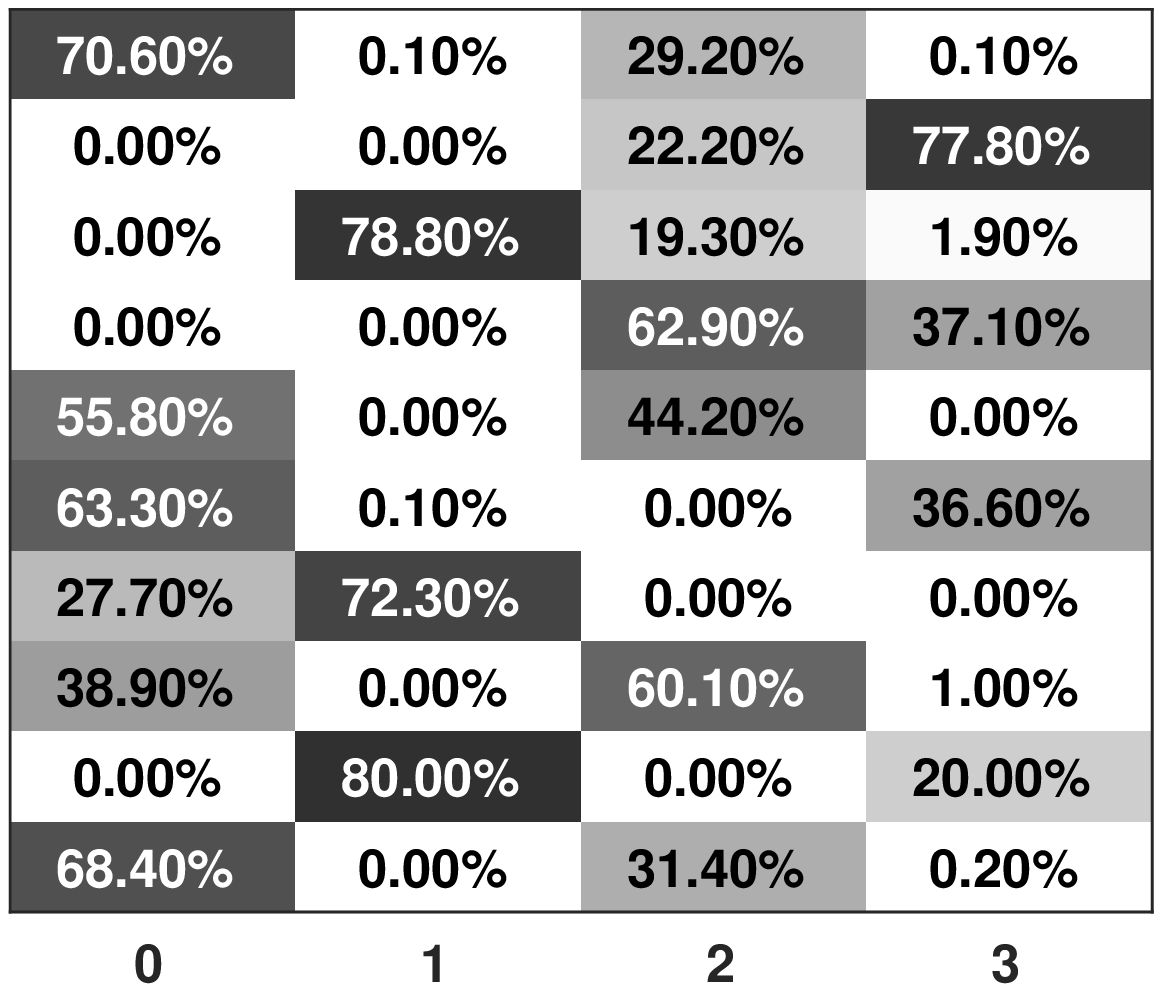}
\caption{k=2}
\end{subfigure}
\begin{subfigure}{0.5\columnwidth}
\centering
\includegraphics[clip=true, trim=40px 10px 35px 20px, scale=0.29]{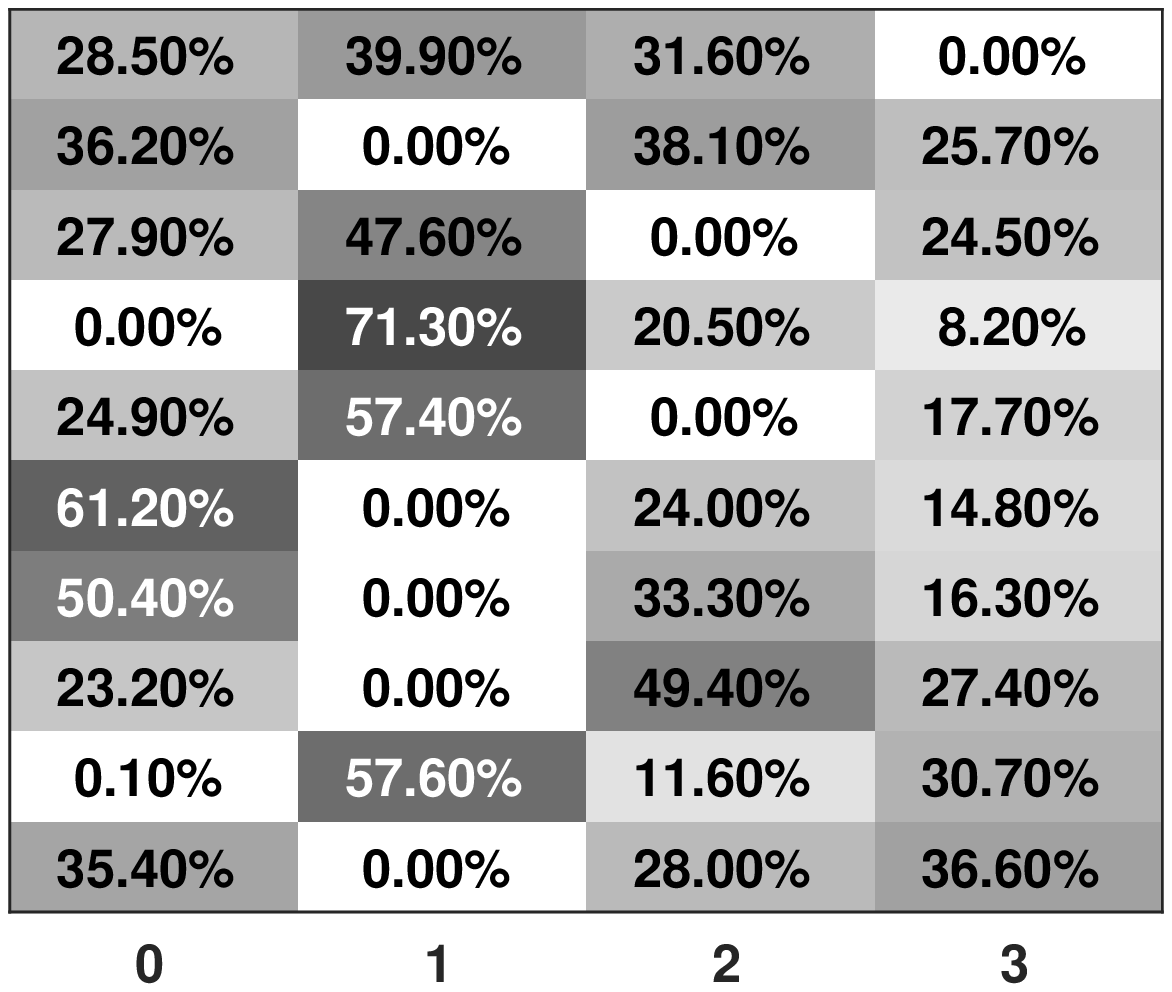}\null\hfill
\caption{k=3}
\end{subfigure}
\begin{subfigure}{0.5\columnwidth}
\centering
\includegraphics[clip=true, trim=40px 10px 35px 20px, scale=0.29]{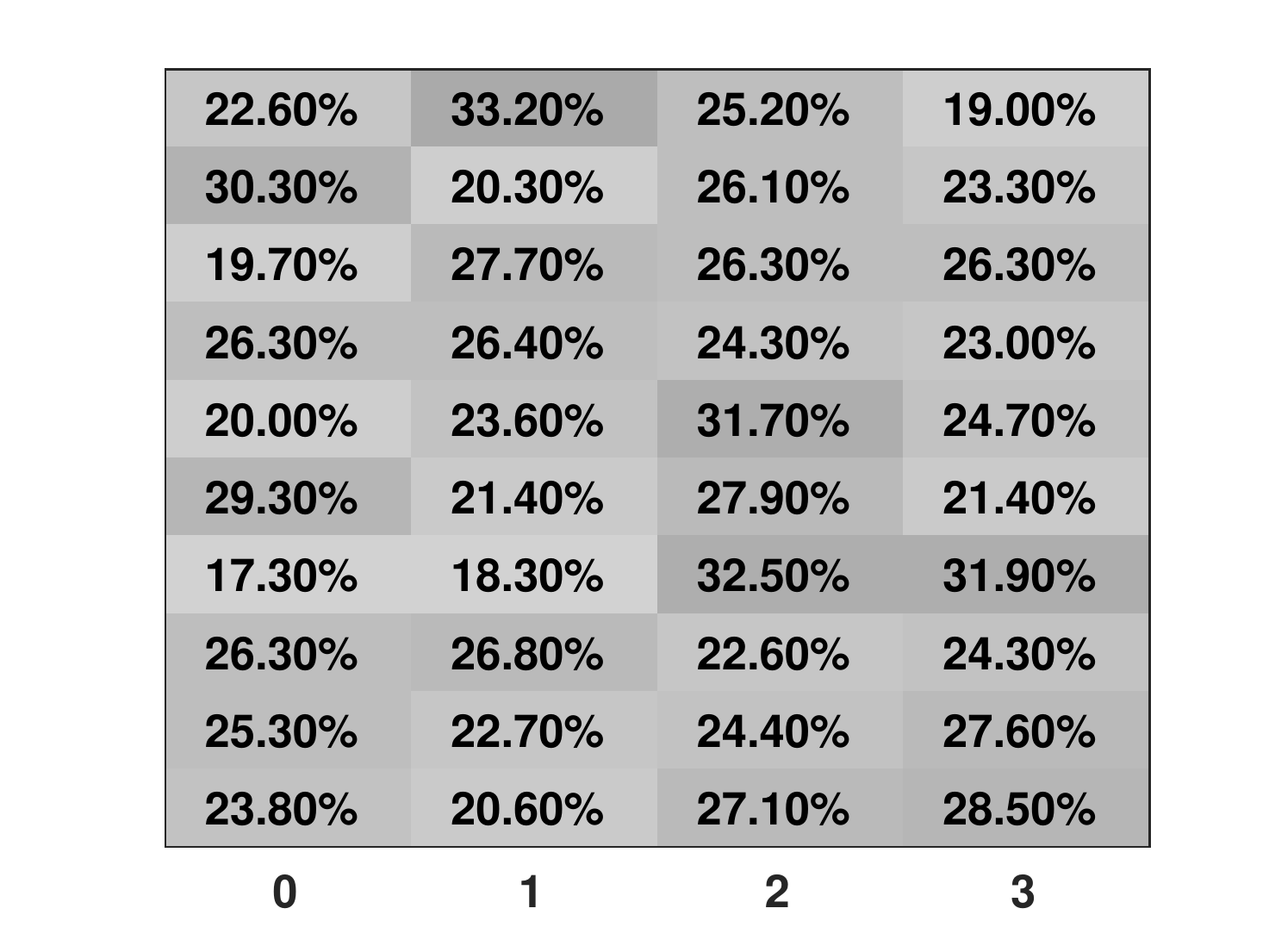}
\caption{k=4}
\end{subfigure}

\begin{subfigure}{2\columnwidth}
\centering
\includegraphics[width=0.9\columnwidth]{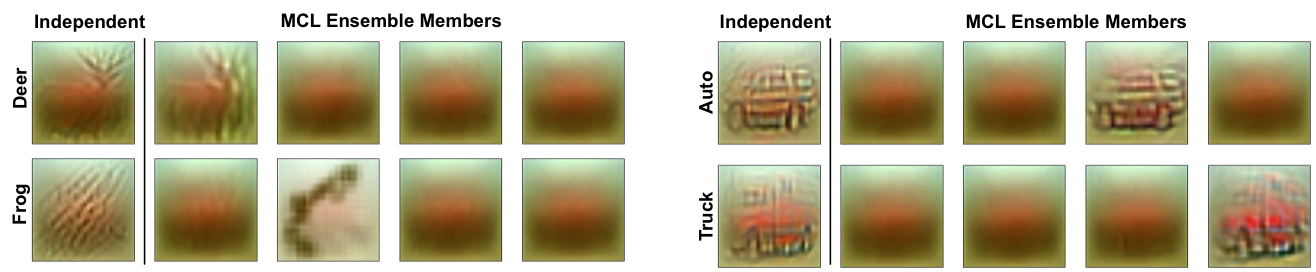}
\caption{}
\end{subfigure}
\caption{\textit{(a)-(d):} Percentage of test examples of each
  class assigned to each ensemble member by the oracle (i.e.\ those with lowest
loss).
The degree of
  specialization is very sharp at $k=1$ and softens to almost uniform
  at $k=4$. \textit{(e):} 
  Guided-backprop images for standard and MCL $k=1$ trained ensemble
  members. Networks that have not specialized in a given
  class are agnostic to the image content.}
\label{fig:vis}
\vspace{-12pt}
\end{figure*}

\xhdr{Experimental results.}
 We begin our experiments with MCL on the CIFAR10-Quick
 network. Table \ref{tab:cifarmcl} shows the individual network
 accuracies and the oracle accuracy for MCL trained ensembles of
 different values of $k$.  As $k$ is increased, each member network is
 exposed to more of the data and we see decreased oracle accuracy in
 exchange for increased individual member performance. At $k$=4, the
 oracle-set loss reduces to independent cross-entropy for each member, producing a standard
 ensemble.  The $k$=1 case showcases the degree of model
 specialization. \textbf{Each individual network performs \emph{very}
 poorly (accuracy of 19-27\%); however, taken as an ensemble the
 oracle accuracy is over 93\%!} This clearly shows that the networks
 have specialized and diversified with each taking responsibility for
 a subset of examples. To the best of our knowledge,
 this is the first work to demonstrate such 
 behavior.

\begin{table}[t]
\centering
\vspace{10pt}
\resizebox{\columnwidth}{!}{
\begin{tabular}{c  c c c c c c}\toprule
\textbf{CIFAR10-Quick}\footnotesize{$\times$4} & \multicolumn{4}{c}{Member Networks} & \multicolumn{2}{c}{Ensemble Accuracy} \\
 k & \multicolumn{4}{c}{Accuracy} & Ensemble-Mean & Oracle \\
\midrule \vspace{-12pt}& & &\\
 1& 24.35  & 27.18 & 27.15 & 19.36 & 28.38 & 93.10  \\
 2& 50.95  & 27.41 & 55.88 & 34.81 & 75.16 & 92.78  \\
 3& 65.46  & 40.71 & 64.79 & 70.37 & 79.76 & 92.55  \\
 4& 77.12  & 76.76 & 77.29 & 76.80 & 80.72 & 89.78  \\\bottomrule
\end{tabular}
}
\vspace{5px}
\caption{Increasing the number of predictors each data point is assigned to results in reduced oracle accuracy as the diversifying effect is reduced. Note that $k=4$ is a standard ensemble.}
\label{tab:cifarmcl}
\vspace{-12pt}
\end{table}

To further characterize what the MCL member networks are learning, we
tracked which test examples are assigned to each ensemble member by
the oracle accuracy metric (i.e. which ensemble member has the lowest
error on each example). Figure \ref{fig:vis}(a)-(d) show the distribution
of classes assigned to each ensemble member, and the results are
striking: at $k$=1 we see almost complete division of the label space!
As $k$ increases we see increased uniformity in these
distributions. Note that these divisions emerge
from the loss and are not hand-designed or pre-initialized in any way.

In Figure~\ref{fig:vis}(e) we visualize how the ensemble members
respond to input images using guided
backprop \cite{SpringenbergDBR14}, which is similar to
the \emph{deconv} visualizations of Zeiler and
Fergus \cite{zeiler2014visualizing}. These images can be interpreted as the gradient of the indicated class score with respect to the input image.
Features that are clear in these images have the largest influence 
on the network's output for a given input image. Each row shows these
visualizations for a single input image for a standard
network and for members of an MCL ensemble.  Networks that have
not specialized in the given class are agnostic to the image
content. See supplementary material for more examples.

\xhdr{MCL As Label-Space Clustering.} We have shown that $k=1$ MCL
trained ensembles tend to converge to a label-space clustering where
each member focuses on a subset of the labels. The set of possible
label-space clusterings is vast, so to put the MCL results into
perspective we train hand-designed specialist ensembles with randomized label assignments. For
CIFAR10 we randomly split the labels evenly to the four ensemble
members and train each with respect to those labels. Over the course
of 100 trials, we found oracle-accuracy ranged from 87.62 to 94.65
with a mean of 91.83. This shows that generally the MCL optimization
selects high quality label space clusterings with respect to oracle
accuracy.

An alternative strategy presented by \cite{hinton2014distilling} is to
diversify members by dividing labels into clusters of hard to
distinguish classes; very briefly described, assignments are
generated by clustering the covariance matrix of label scores computed
across an input set for a generalist CNN. We trained an ensemble using this clustering method and it led to significantly decreased oracle performance versus MCL on CIFAR10-Quick and ILSVRC-Alex. This is not surprising since they do not optimize for oracle accuracy.

\xhdr{Overcoming Data Fragmentation.}  Despite not training member
networks to convergence in each iteration of coordinate descent, our
method results in improved oracle accuracy over standard
ensembles. However, interleaving the assignment step with stochastic
gradient descent results in data fragmentation, with each network
seeing only a fraction of each batch (as illustrated by the
class-specialization). We find this reduced effective batch size results in
noisy gradients that inhibit learning, especially on larger networks.

Deep networks are especially sensitive to the effects of data
fragmentation early in training when errors (and therefore gradients)
are typically larger. In Guzman-Rivera et al.~\cite{rivera_nips12},
initial assignments for the first iteration of training were decided
by clustering the data into $k$ clusters. In contrast, assignments in
our approach are based on network performance which is initially the
result of random initialization. To investigate the effect of this
initial phase of learning, we applied our MCL loss to fine-tune a
previously trained CIFAR10-Quick ensemble. As shown in Table
\ref{tab:pretrain}, the benefits of pretraining are most pronounced
for lower values of $k$ where data fragmentation is most severe.

\begin{table}[t]
\centering
\resizebox{0.9\columnwidth}{!}{
\begin{tabular}{c  c c c c c}\toprule
\textbf{CIFAR10-Quick}\footnotesize{$\times$4} & \multicolumn{5}{c}{ Iterations of Cross-Entropy Pretraining} \\
 k  & 100 & 500 & 1000 & 2000 & 4000 \\
\midrule \vspace{-12pt}& & &\\
 1&  94.21  & 94.21 & 94.65 & 95.75 & 96.00  \\
 2&  92.79  & 93.06 & 93.16 & 93.07 & 93.00 \\
 3&  92.25  & 92.93 & 91.77 & 90.94 & 90.94 \\\bottomrule
\end{tabular}
}
\vspace{5px}
\caption{Increasing the amount of pretraining before fine-tuning with the MCL loss results in increase oracle accuracy.}
\label{tab:pretrain}
\vspace{-10pt}
\end{table}

While pretraining did stabilize learning, data fragmentation on
CIFAR10 is a relatively minor problem whereas training with MCL from scratch
on larger networks using standard batch sizes consistently failed to
outperform standard ensembles. We attribute this to a combination of
data fragmentation and the difficulty of initial learning.  To test
this hypothesis, we experimented with fine-tuning and gradient
accumulation across batches on the ILSVRC-Alex architecture. We accumulated gradients from 5 batches before updating
parameters and fine-tuned from a fully-trained
ensemble. Table \ref{tab:alexmcl} shows the result of 3000 iterations
of this fine-tuning experiment for different values of $k$. This setup
overcame the data fragmentation problem and we see the same trends as
in CIFAR10.

These experiments demonstrate MCL's ability to quickly diversify an
ensemble. To push this further, we reran the fine-tuning experiment
for $k$=1, this time initializing all ensemble members with the
same network. Despite starting from an ensemble of identical networks
with an oracle accuracy of 56.90\%, the ensemble reached an oracle
accuracy of 72.67\% after only 3000 iterations!

\begin{table}[t]
\centering
\resizebox{0.95\columnwidth}{!}{
\begin{tabular}{c c c c c}\toprule
\textbf{ILSVRC-Alex}\footnotesize{$\times$5} & \multicolumn{2}{c}{Single Member} & \multicolumn{2}{c}{Ensemble Accuracy} \\
  & \multicolumn{2}{c}{Accuracy} &  Ensemble-Mean & Oracle \\
\midrule \vspace{-12pt}& & &\\
 k=1 & \multicolumn{2}{c}{46.50} &  55.22 & 74.67 \\
 k=2 & \multicolumn{2}{c}{52.48} &  59.21 & 73.40 \\
 k=3 & \multicolumn{2}{c}{55.38} &  59.73 & 71.75 \\
 k=4 & \multicolumn{2}{c}{56.33} &  60.09 & 70.84 \\
 base ensemble & \multicolumn{2}{c}{57.17} &  60.31 & 70.50 \\
 \bottomrule
\end{tabular}
}
\vspace{5px}
\caption{Fine-tuning and gradient accumulation across batches allows larger networks to specialized under the MCL loss.}
\label{tab:alexmcl}
\vspace{-15pt}
\end{table}

We have demonstrated that the MCL loss is effective at inducing
diversity, however the member networks specialize so much
that Ensemble-Mean Accuracy suffers. We tried
linearly combining the MCL loss with the standard cross-entropy to
balance diversity with general performance. We find training under this loss 
improves CIFAR10 Ensemble-Mean accuracy by 1\% over a standard
ensemble.

In this section we have developed a novel MCL framework and shown it produces ensembles with substantially improved oracle accuracies when training from scratch and even when fine-tuning from a single network.




\section{Distributed Ensemble Training}
\label{sec:mpi}
\begin{figure}[t]
\centering
\includegraphics[trim=0px -10px 16px 0px, clip=true, width=0.91\columnwidth]{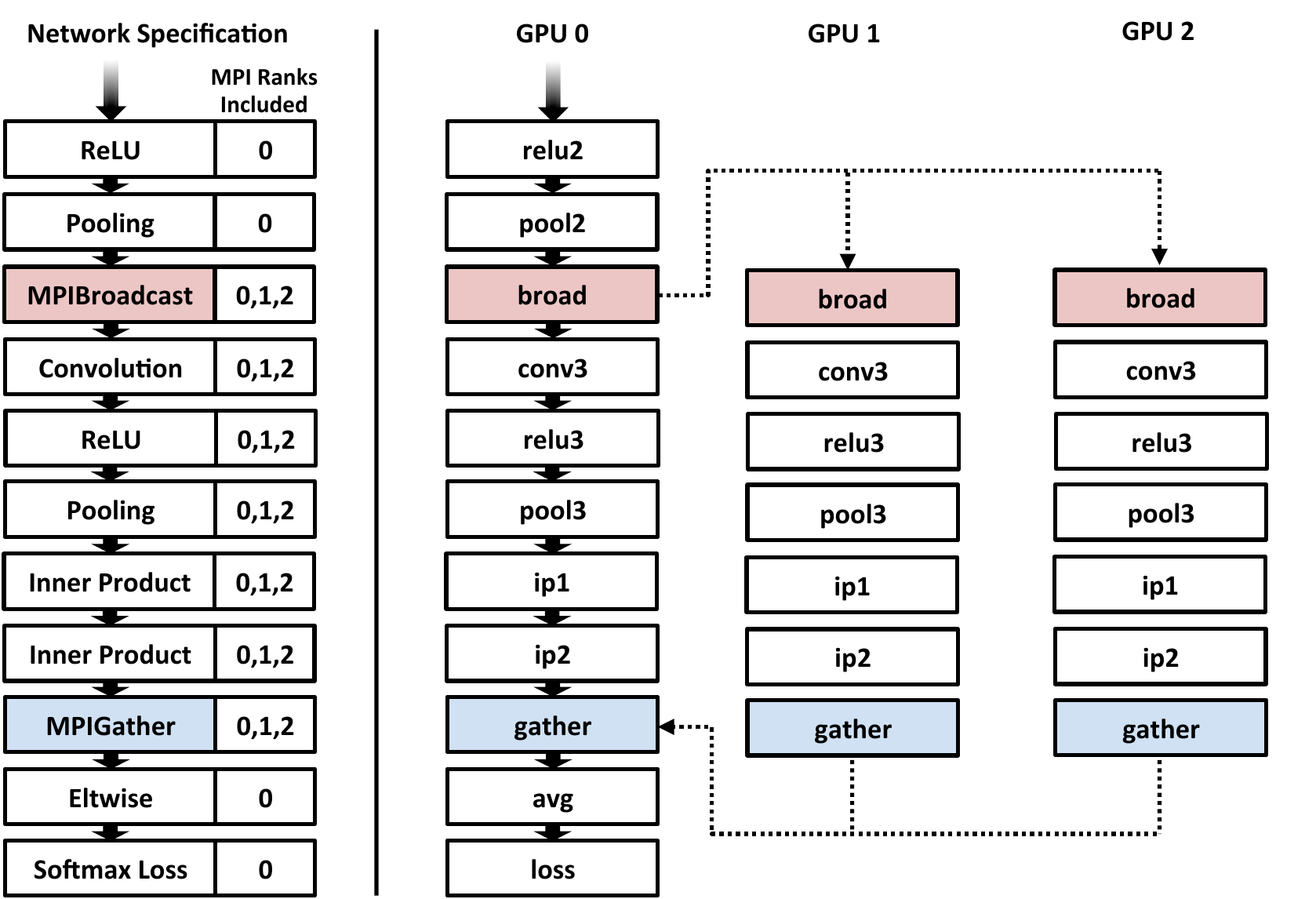}
\caption{MPI-Caffe models can be defined by a single network specification and distributed by across multiple GPUs. Dashed lines indicate cross-process communication and not input/output.}
\label{fig:parse}
\vspace{-10pt}
\end{figure}

Training an ensemble on a single GPU is prohibitively expensive,
so standard practice for large ensembles is to train
the multiple networks either sequentially or in parallel.
However, any form of model coupling requires communication
between learners. To make enable our experiments at scale, we have developed and will release a
modification to Caffe, which we call MPI-Caffe, that uses the Message
Passing Interface (MPI)~\cite{mpi3_1} standard to enable
cross-GPU/machine communication.
 These communication operations are provided as Caffe model
layers, allowing network designers to quickly
experiment with distributed networks, where different parts of the model reside on different GPUs and machines.  Figure~\ref{fig:parse} shows how
an ensemble of CIFAR10-Quick networks with parameter sharing and model
averaging is defined as a single specification and distributed
across multiple process. In MPI-Caffe, each process is assigned a identifier (called a rank); by setting the ranks each network layer belongs
to, we can easily design distributed ensembles.

The {\small{\texttt{MPIBroadcast}}} and {\small{\texttt{MPIGather}}} layers provide the
core communication functionality.  {\small{\texttt{MPIBroadcast}}} forwards
its input to the other processes during a forward pass and accumulates
gradients from each during back-propagation. The forward pass for 
{\small{\texttt{MPIGather}}} collects all of its inputs from multiple
processes and outputs them to a single network, and the backward pass
 simply routes the gradients back to the
corresponding input.

We tested our MPI-Caffe framework on a large-scale cluster with one Telsa K20
GPU per node and a maximum MPI node interconnect bandwidth of 5.8 GB/sec. 
To characterize the communication overhead for an
ensemble, we measure the time spent sharing various layers of the
ILSVRC-Alex{\footnotesize $\times$5} architecture. The largest layer
we shared was \emph{pool2} which amounts to broadcasting nearly 36
million floats per batch. Despite the
layer's size we find only 0.49\% of the forward-backward pass time is
used by communication. More details are available in the
supplement.

\section{Discussion and Conclusion}
\label{sec:conclusions}

There is a running theme behind all of the ideas presented in this
paper: diversity. Our experiments on bagging demonstrate
that the diversity induced in ensemble members by random parameter
initializations is more useful than that introduced by bags with
duplicated examples. Our experiment on explicitly training for
Ensemble-Mean performance show that averaging beliefs of ensemble
members before computing losses has the unintended effect of
removing diversity in gradients. Our novel diversity-inducing MCL loss
shows that encouraging diversity in ensemble members can significantly
improve performance.  Finally, our novel TreeNet architecture shows
that diversity is important in high-level representations while
low-level filters might be better off without it.
Training these large-scale architectures is made practical by our
MPI-Caffe framework.


In future work, we would like to adapt the MCL
framework to structured predictions problems. In a
structured context where the space of ``good'' solutions is quite
large, we feel diverse models can have an even greater
benefit.

{\footnotesize
\bibliographystyle{ieee}
\bibliography{strings,dbatra}
}

\begin{SaveVerbatim}[commandchars=&\[\]]{BROAD}
1 layer{ 
&fvtextcolor[red][2  name: broad ]
&fvtextcolor[red][3  type: MPIBroadcast]
&fvtextcolor[red][4  bottom: pool2 ]
&fvtextcolor[red][5  top: pool2_b] 
&fvtextcolor[darkorchid][6  mpi_param{ ]
&fvtextcolor[darkorchid][7    root: 0 ]
&fvtextcolor[darkorchid][8  }]
&fvtextcolor[blue][9  include{]
&fvtextcolor[blue][10    mpi_rank: 0]
&fvtextcolor[blue][11    mpi_rank: 1]
&fvtextcolor[blue][12    mpi_rank: 2]
&fvtextcolor[blue][13  }]
14 }
\end{SaveVerbatim}

 \newpage
 \onecolumn


\appendix
\gdef\thesection{Appendix \Alph{section}}

\newpage
\section{TreeNet Object Detection Results on PASCAL VOC 2007 }\label{app:pascal}
As briefly described in Section 5 of the main paper, the ILSVRC-Alex
TreeNet architecture was also evaluated for object detection using the
PASCAL VOC 2007 dataset, which includes labeled ground-truth
bounding-box annotations for 20 object classes. For this task, we used
Fast R-CNNs~\cite{girshick2015fast}. During training, Fast RCNNs
finetune a model pretrained on ImageNet for classification under two
losses, one over the predicted class of an object proposal, and
one with bounding box regression. For our ensembles, we average both the class
prediction as well as the bounding box coordinates from ensemble
member models.

To evaluate TreeNets and standard ensembles, we fine-tune four
different instances for each under the Fast R-CNN framework and
compute the mean and standard deviation of the classwise average
precisions (APs) as well as the mean APs over all classes. Table
\ref{tab:frcnn_bbox} presents these results for various models with
the averaged bounding boxes -- a standard ensemble, a TreeNets split
after conv1, conv2, and conv3, as well as a single model. We remind
the reader that non-parameterized layers are irrelevant with respect
to splits so we do not report results for those layers. We also
evaluate without the bounding-box regression, instead using the initial
selective search proposals directly. Table \ref{tab:frcnn_nobbox}
shows these results.

In both tasks we see that TreeNets outperform the standard ensembles
and single models by significant margins. We note that we see similar
gains in accuracy when using the regressed bounding boxes for both
single models and ensembles, implying that the bounding box averaging
procedure for ensembles is reasonable.

\vspace{20pt}
\def\LineSpace{1.5ex}
\def\StdevSpace{-0.8ex}
\begin{minipage}{0.95\textwidth}
\vspace{8pt}
\begin{center}

\setlength{\tabcolsep}{4pt}
\resizebox{\textwidth}{!}{
\begin{tabular}{c  c  c   c   c  c  c  c  c  c  c  c  c  c  c  c  c  c  c  c  c | c@{}}

\toprule
$\mu\pm\sigma$ & \includegraphics[width=20px]{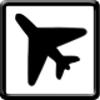} & \includegraphics[width=20px]{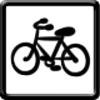} & \includegraphics[width=20px]{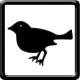} & \includegraphics[width=20px]{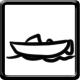} & \includegraphics[width=20px]{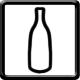} & \includegraphics[width=20px]{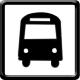} & \includegraphics[width=20px]{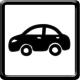} & \includegraphics[width=20px]{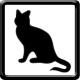} & \includegraphics[width=20px]{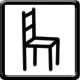} & \includegraphics[width=20px]{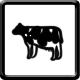} & \includegraphics[width=20px]{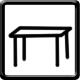} & \includegraphics[width=20px]{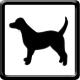} & \includegraphics[width=20px]{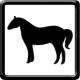} & \includegraphics[width=20px]{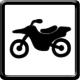} & \includegraphics[width=20px]{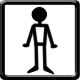} & \includegraphics[width=20px]{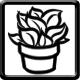} & \includegraphics[width=20px]{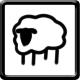} & \includegraphics[width=20px]{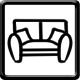} & \includegraphics[width=20px]{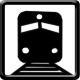} & \includegraphics[width=20px]{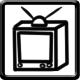} & mean\\
\midrule
\multirow{2}{*}{Ensemble} & 67.49 & 68.89 & 52.00 & 38.25 & 16.74 & 68.14 & 70.68 & 67.56 & 26.45 & 63.78 & 61.93 & 61.74 & 73.31 & 67.18 & 56.53 & 23.84 & 50.45 & 54.81 & 69.04 & 59.26 & 55.90 \\[\StdevSpace]
& \spm1.42 & \spm0.57 & \spm1.25 & \spm2.38 & \spm0.90 & \spm1.90 & \spm0.37 & \spm2.65 & \spm0.16 & \spm1.63 & \spm1.59 & \spm1.69 & \spm0.68 & \spm0.56 & \spm0.30 & \spm0.58 & \spm0.82 & \spm1.45 & \spm1.45 & \spm0.88 & \spm0.21\\[\LineSpace]

\multirow{2}{*}{conv1} & \bf 67.83 & 68.47 & {\bf52.66} & \bf37.90 & \bf 18.01 & {\bf69.53} & 70.83 & \bf67.93 & {\bf27.30} & 61.59 & \bf62.96 & 62.06 & {\bf74.89} & 67.97 & 57.43 & 23.80 & 50.94 & {\bf56.42} & 70.57 & 59.79 & 56.44 \\[\StdevSpace]
& \spm1.34 & \spm0.85 & \spm1.41 & \spm0.98 & \spm0.81 & \spm1.85 & \spm0.51 & \spm0.49 & \spm0.84 & \spm2.31 & \spm0.65 & \spm1.16 & \spm1.14 & \spm0.48 & \spm0.33 & \spm0.50 & \spm2.06 & \spm0.55 & \spm1.53 & \spm0.91 & \spm0.23\\[\LineSpace]

\multirow{2}{*}{conv2} & 67.30 & {\bf69.29} & 52.62 & 37.58 & 17.90 & 68.81 & \bf71.04 & 68.54 & 27.13 & \bf63.66 & 62.37 & {\bf62.20} & 74.60 & \bf68.45 & {\bf57.52} & \bf24.20 & {\bf52.53} & 55.00 & \bf71.31 & {\bf60.66} & {\bf56.64} \\[\StdevSpace]
& \spm1.05 & \spm0.79 & \spm1.41 & \spm1.41 & \spm0.82 & \spm0.73 & \spm0.28 & \spm1.50 & \spm1.05 & \spm1.06 & \spm1.83 & \spm0.62 & \spm0.56 & \spm0.23 & \spm0.46 & \spm1.24 & \spm0.59 & \spm0.73 & \spm1.64 & \spm0.19 & \spm0.23\\[\LineSpace]

\multirow{2}{*}{conv3} & 66.29 & 67.35 & 50.22 & 36.23 & 16.65 & 67.77 & 70.22 & 67.73 & 25.01 & 60.91 & 61.80 & 61.99 & 73.64 & 68.38 & 56.42 & 21.57 & 49.75 & 55.54 & 70.03 & 58.81 & 55.32 \\[\StdevSpace]
& \spm1.49 & \spm1.05 & \spm1.40 & \spm2.54 & \spm0.83 & \spm2.21 & \spm0.73 & \spm0.48 & \spm1.62 & \spm3.27 & \spm1.03 & \spm0.69 & \spm0.51 & \spm0.92 & \spm1.02 & \spm0.42 & \spm1.95 & \spm1.26 & \spm0.77 & \spm0.58 & \spm0.54\\[\LineSpace]

Single  &  62.53  &  65.25  &  41.30  &  32.36  &  11.98  &  62.56  &  66.89  &  61.76  &  20.72  &  56.16  &  56.91  &  55.14  &  69.53  &  64.48  &  51.39  &  21.06  &  45.20  &  47.91  &  65.70  &  54.35  &  50.66\\[\StdevSpace]
  Model &  \spm0.24  &  \spm0.49  &  \spm2.12  &  \spm2.47  &  \spm1.11  &  \spm1.27  &  \spm0.80  &  \spm2.14  &  \spm0.58  &  \spm3.32  &  \spm1.58  &  \spm1.56  &  \spm1.71  &  \spm1.27  &  \spm1.10  &  \spm0.36  &  \spm0.77  &  \spm1.44  &  \spm0.89  &  \spm1.94  &  \spm0.30\\
\bottomrule
\end{tabular}
}
\end{center}
\vspace{-10pt}
\captionof{table}{Average Precision for Object Detection using different TreeNet models with the Fast R-CNN framework, when the coordinates of bounding boxes from each member model are averaged.}
\label{tab:frcnn_bbox}
\end{minipage}

\vspace{50pt}
\begin{minipage}{0.95\textwidth}
\vspace{8pt}
\begin{center}
\setlength{\tabcolsep}{4pt}
\resizebox{\textwidth}{!}{
\begin{tabular}{c  c   c c   c  c  c  c  c  c  c  c  c  c  c  c  c  c  c  c  c | c@{}}
\toprule
$\mu\pm\sigma$ & \includegraphics[width=20px]{images/airplane.jpg} & \includegraphics[width=20px]{images/bike.jpg} & \includegraphics[width=20px]{images/bird.jpg} & \includegraphics[width=20px]{images/boat.jpg} & \includegraphics[width=20px]{images/bottle.jpg} & \includegraphics[width=20px]{images/bus.jpg} & \includegraphics[width=20px]{images/car.jpg} & \includegraphics[width=20px]{images/cat.jpg} & \includegraphics[width=20px]{images/chair.jpg} & \includegraphics[width=20px]{images/cow.jpg} & \includegraphics[width=20px]{images/table.jpg} & \includegraphics[width=20px]{images/dog.jpg} & \includegraphics[width=20px]{images/horse.jpg} & \includegraphics[width=20px]{images/motorbike.jpg} & \includegraphics[width=20px]{images/person.jpg} & \includegraphics[width=20px]{images/plant.jpg} & \includegraphics[width=20px]{images/sheep.jpg} & \includegraphics[width=20px]{images/sofa.jpg} & \includegraphics[width=20px]{images/train.jpg} & \includegraphics[width=20px]{images/tv.jpg} & mean\\
\midrule
\multirow{2}{*}{Ensemble} & \bf64.48 & 63.99 & 45.42 & 34.17 & 15.60 & 63.90 & 67.81 & 62.55 & 24.20 & \bf58.36 & 55.91 & {\bf56.81} & 62.61 & {\bf67.76} & 50.35 & 22.25 & 45.79 & 49.47 & 65.20 & 57.76 & 51.72  \\[\StdevSpace]
 & \spm0.75 & \spm0.94 & \spm{0.71} & \spm1.14 & \spm0.26 & \spm1.46 & \spm0.34 & \spm0.90 & \spm0.27 & \spm2.09 & \spm1.34 & \spm1.14 & \spm1.24 & \spm0.52 & \spm0.35 & \spm0.84 & \spm1.68 & \spm1.76 & \spm1.57 & \spm0.72 & \spm0.15  \\[\LineSpace]
\multirow{2}{*}{conv1} & 64.32 & {\bf64.80} & {\bf46.45} & \bf36.04 & {\bf17.03} & 65.61 & 68.24 & 63.21 & {\bf25.06} & 56.79 & 56.90 & 56.80 & \bf62.32 & 66.58 & 50.68 & {\bf22.68} & 46.66 & \bf49.76 & 66.03 & 58.24 & {\bf 52.21}  \\[\StdevSpace]
 & \spm1.01 & \spm0.73 & \spm1.04 & \spm1.69 & \spm1.02 & \spm1.43 & \spm0.88 & \spm1.43 & \spm0.36 & \spm2.57 & \spm0.99 & \spm1.33 & \spm1.90 & \spm1.31 & \spm0.38 & \spm0.81 & \spm1.99 & \spm0.97 & \spm0.75 & \spm1.53 & \spm0.42  \\[\LineSpace]
\multirow{2}{*}{conv2} & 64.22 & 63.68 & 45.85 & 35.31 & 16.43 & {\bf65.89} & {\bf68.78} & \bf63.21 & 24.57 & 56.37 & 57.10 & 56.57 & 62.28 & 66.69 & {\bf51.04} & 21.72 & {\bf48.19} & 48.86 & 66.24 & {\bf58.71} & 52.08  \\[\StdevSpace]
 & \spm1.04 & \spm1.51 & \spm1.24 & \spm0.63 & \spm0.70 & \spm1.63 & \spm0.25 & \spm1.60 & \spm0.86 & \spm1.40 & \spm1.50 & \spm1.28 & \spm0.77 & \spm1.49 & \spm0.56 & \spm0.96 & \spm0.80 & \spm1.31 & \spm2.19 & \spm1.06 & \spm0.33  \\[\LineSpace]
\multirow{2}{*}{conv3} & 63.56 & 63.91 & 44.18 & 34.24 & 16.01 & 64.36 & 67.71 & 63.00 & 23.27 & 56.99 & \bf57.21 & 56.31 & 61.05 & 66.04 & 50.22 & 20.22 & 45.03 & 49.40 & {\bf66.26} & 56.92 & 51.29  \\[\StdevSpace]
 & \spm1.10 & \spm0.75 & \spm0.67 & \spm2.78 & \spm0.96 & \spm1.57 & \spm0.86 & \spm1.41 & \spm0.82 & \spm1.71 & \spm0.46 & \spm1.49 & \spm1.65 & \spm1.18 & \spm0.85 & \spm0.29 & \spm2.39 & \spm0.73 & \spm1.80 & \spm0.56 & \spm0.44  \\[\LineSpace]
Single & 59.11 & 60.82 & 35.84 & 28.84 & 10.73 & 58.91 & 63.98 & 55.84 & 19.40 & 50.03 & 51.65 & 47.95 & 58.38 & 61.44 & 44.81 & 18.61 & 41.64 & 42.74 & 61.73 & 51.89 & 46.22  \\[\StdevSpace]
 Model& \spm1.49 & \spm1.04 & \spm1.86 & \spm1.20 & \spm0.86 & \spm2.04 & \spm0.69 & \spm1.31 & \spm0.78 & \spm2.91 & \spm2.91 & \spm2.48 & \spm2.49 & \spm1.06 & \spm0.99 & \spm0.38 & \spm2.17 & \spm1.32 & \spm1.38 & \spm0.63 & \spm0.26  \\
\bottomrule
\end{tabular}
}
\end{center}
\vspace{-10pt}
\captionof{table}{Average Precision for Object Detection using different TreeNet models with the Fast R-CNN framework, when predicted bounding boxes are not used.}
\label{tab:frcnn_nobbox}
\end{minipage}
\newpage
\section{Instability of Averaged Softmax Outputs }\label{app:avg}
\noindent As discussed in Section 6.1 of the main paper, training under a cross-entropy loss over averaged softmax outputs results in reduced performance compared to both standard ensembles and score-averaged ensembles. We find that this is because averaging softmax outputs prior to the cross-entropy loss has less stable gradients compared to standard cross-entropy over softmax outputs. Let us consider the standard case first and formulate the derivative of the cross-entropy loss with respect to softmax inputs. The cross-entropy loss and softmax function are defined as:
$$\ell(x,y) = -\log(p_y) \mbox{\hspace{10pt}and\hspace{10pt}} p_y = \frac{e^{s_y}}{\sum_{i} e^{s_i}}$$
The derivative of the softmax probability $p_y$ with respect to some score $s_j$ is 
$$\frac{\partial p_y}{\partial s_j} = p_y(I_{jy}-p_j)$$
where $I_{jy}$ is 1 if $y=j$ and 0 otherwise. This derivative  requires multiplying probabilities which can be quite small, leading to underflow errors. Taking the derivative of the cross-entropy loss with respect to some $s_j$ results in a more stable solution:
$$\frac{\partial\ell}{\partial s_j} = \frac{\partial \ell}{\partial p_y} \frac{\partial p_y}{\partial s_j} = -\frac{1}{p_y} p_y(I_{jy}-p_j) = p_j - I_{jy}$$
Let us now consider the case where $p_y$ is averaged over $M$ predictors such that
$$p_y = \frac{1}{M} \sum_m p_y^m \mbox{\hspace{10pt}and\hspace{10pt}} p_y^m = \frac{e^{s_y^m}}{\sum_i e^{s_i^m}}$$
The derivative of this new $p_y$ with respect to the score of one predictor $s_j^m$ is then
$$\frac{\partial p_y}{\partial s_j^m} = \frac{\partial p_y}{\partial p_y^m} \frac{\partial p_y^m}{\partial s_j^m} = \frac{1}{M}\frac{\partial p_y^m}{\partial s_j^m}= \frac{1}{M} p_y^m(I_{jy}-p_j^m)$$
Again computing the derivative of the loss with respect to a score $s_j^m$ we see
$$\frac{\partial \ell}{\partial s_j^m} = \frac{\partial \ell}{\partial p_y} \frac{\partial p_y}{\partial p_y^m} \frac{\partial p_y^m}{\partial s_j^m} = \frac{1}{p_y}\frac{1}{M} p_y^m(p_j^m-I_{jy}) = \frac{p_y^m}{\sum_i p_y^i}  (p_j^m -I_{jy})$$

\noindent The rightmost term in this result is identical to the standard case presented above; however, the first term acts to weight the gradient for each predictor and can be shown to range from 0 to M.  The product of this term and the probability $p_j^m$ can be prone to underflow errors when $p_y^m$ is less than $p_y$.  On the other hand, when $p_y^m$ is greater than $p_y$ the gradients are increased in magnitude which can result in overshooting good minima.

This scaling of the gradients has an interesting similarity with MCL. If a predictor puts little mass into the correct class compared to the other predictors, the weighting factor and thus the gradient go to zero -- meaning worse performing members are less encouraged to improve than strong performers.  This is similar behavior to what a soft-assignment variant of MCL might induce. However, we do not notice improved oracle accuracy relative to base ensembles for models trained with probability-averaged losses, implying the predictors are making relatively similar predictions.

\newpage
\section{Pseudo-code for Stochastic Gradient Descent with MCL}\label{app:sgdmcl}
\noindent We describe the classical MCL algorithm and our approach to integrate MCL coordinate descent with stochastic gradient descent in Section 6.2 of the main paper. Here we provide psuedocode for both algorithms to highlight the differences and provide additional clarity. 

\begin{minipage}{\textwidth}
\vspace{10pt}
\begin{algorithm}[H]
 \KwData{$\mbox{Dataset }(x_i, y_i) \in D$ and loss $\mathcal{L}$}
 \KwResult{ Predictor parameters $\theta_1 \cdots \theta_M$ }
  \textbf{Initialization:}\\
  \hspace{5pt}$D^*_0=\{D_1 \cdots D_M\} \leftarrow \mbox{k-means}(D, k=M)$\\
  \hspace{5pt}$t \leftarrow 0$\\
 \While{$D^*_{t} \neq D^*_{t-1}$}{
  \textbf{Step 1: Train each predictor to completion using only its corresponding subset of the data}\\
  \hspace{10pt}$\theta_m \leftarrow \mbox{Train}(D_m)$\\
  \textbf{Step 2: Reassign each example to its least-loss predictor}\\
  \hspace{10pt} $D_m = \{ d\in D | \forall \theta_j, \mathcal{L}(d;\theta_m) \leq  \mathcal{L}(d;\theta_j)\}$\\
  $t \leftarrow t+1$
 }
 \caption{Classical MCL}
\end{algorithm}
\end{minipage}

\vspace{20pt}

\begin{minipage}{\textwidth}
\begin{algorithm}[H]
 \KwData{$\mbox{Dataset }(x_i, y_i) \in D$, SGD parameters $\eta,\lambda$, and loss $\mathcal{L}=\sum_i \ell$}
 \KwResult{ Network parameters $\theta_1 \cdots \theta_M$ }
 \textbf{Initialization:}\\
 \hspace{5pt}Randomly initialize $\theta_1 \cdots \theta_M$\\
 \hspace{5pt}$t \leftarrow 0$\\
 \While{$\mathcal{L}_{set}^{t}  < \mathcal{L}_{set}^{t-1}$}{
 $t \leftarrow t+1$\\
   Sample batch $B \subset D$\\
   \textbf{Step 1: Forward pass}\\
  \hspace{5pt} For $b \in B$, compute forward-pass and  losses $\ell(b;\theta_1) \cdots \ell(b;\theta_M)$\\
  \hspace{5pt} Partition $B$ by updating indicator variables $\alpha$ as:\\
   \hspace{15pt}$\alpha_{mi} = 1[[ m = argmin_{m' in 1:M} \ell(b; theta_m')]]$\\
  \hspace{5pt} $\mathcal{L}_m = \sum_i \alpha_{mi}\ell(b_i; \theta_m)$\\
  \hspace{5pt} $\mathcal{L}_{set}^t = \sum \mathcal{L}_m$\\
  \textbf{Step 2: Backward pass}\\
  \hspace{5pt} For each $\theta_m$ apply gradient descent update using only the subset of examples on which it achieves the lowest loss\\
  \hspace{10pt}$\theta_m \leftarrow \theta_m -\eta\nabla\mathcal{L}_m - \lambda\Delta\theta_m$\\
  
 }
 \caption{Integrating MCL coordinate descent with SGD steps}
\end{algorithm}
\end{minipage}
\newpage
\section{Visualizations for MCL Trained Ensembles}\label{app:vis}

\noindent In this section we present additional insight into how MCL ensemble training differs from the behavior of standard ensembles. To show how the distribution of class examples changes over training for MCL we have produced a video showing the proportion of each CIFAR10 class assigned to each predictor at test time and how it changes over training iterations. The intensity of each class icon is proportional to the fraction of class examples assigned to a predictor.  Figure \ref{fig:movie} shows a sample early and later frame from the video.

\begin{minipage}{0.95\textwidth}
\vspace{10pt}
\center
\includegraphics[width=0.35\textwidth]{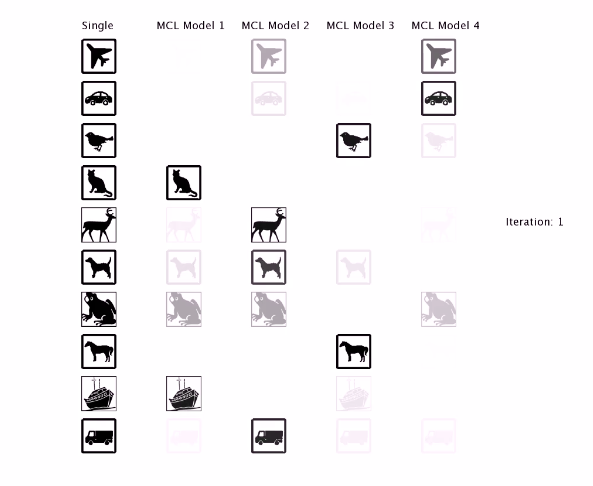}\hspace{20pt}
\includegraphics[width=0.35\textwidth]{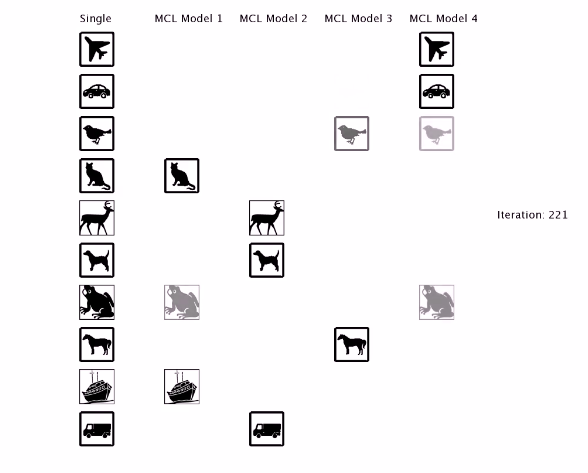}\\
\captionof{figure}{The left image shows the class distribution early in training -- notice how many classes are split between multiple predictors. The right frame shows the evolution of this distribution after 220 additional iterations. Many of the classes have stabilized.}
\label{fig:movie}
\vspace{10pt}
\end{minipage}

We also present additional  guided-backprop \cite{SpringenbergDBR14} visualizations described in Section 6.2 of the main paper for different layers in members of traditional and MCL ensembles.  These images visualize how the ensemble members respond to input images. These images can be interpreted as the gradient of a neuron output with respect to the input image. Features that are clear in these images have the largest influence 
on the network's output for a given input image. Figure~\ref{fig:ip2_vis} shows these visualizations taken for an input image with respect to its true class label. Notice that ensemble members are agnostic to classes that they are not specialized in. The input images are those that produce the highest correct response on the ensemble model.
Visualisations of the same neurons in Figure~\ref{fig:ip2_vis_indep} are generated independently for each model using the image that gives the highest activation. We note that while there is a greater response for non-specialized ensemble members, they remain largely indifferent to image content. We see similar patterns of indifference in lower convolutional layer visualizations as well shown in Figures \ref{fig:conv1_vis}, \ref{fig:conv2_vis}, and \ref{fig:conv3_vis}.

\begin{minipage}{0.95\textwidth}
\centering
\includegraphics[width=0.4\textwidth]{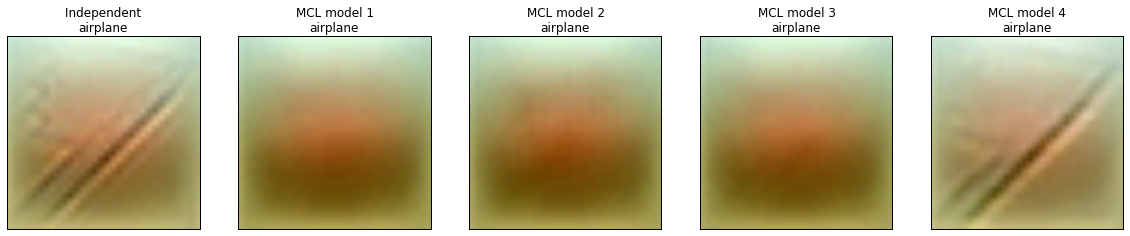}\hspace{20pt}
\includegraphics[width=0.4\textwidth]{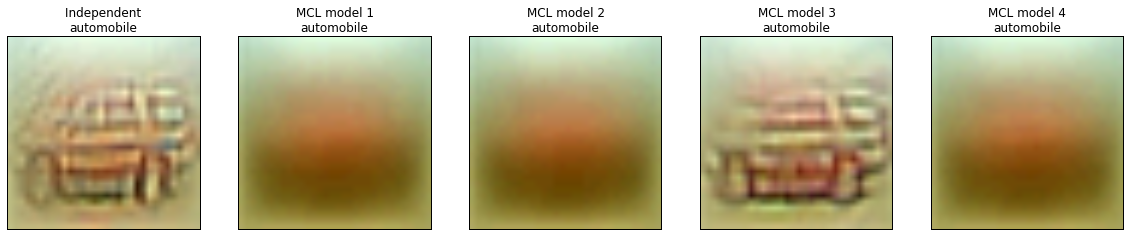}\\
\includegraphics[width=0.4\textwidth]{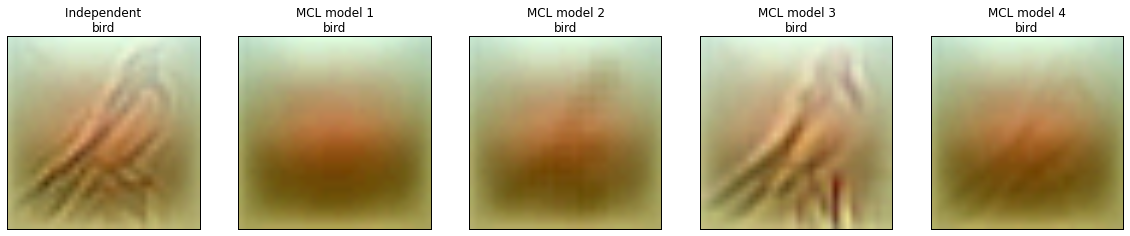}\hspace{20pt}
\includegraphics[width=0.4\textwidth]{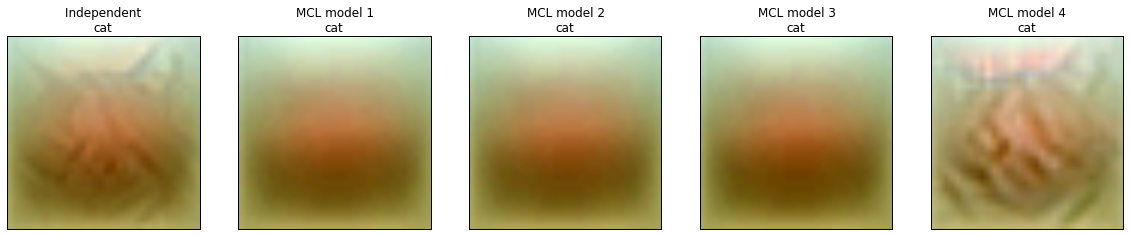}\\
\includegraphics[width=0.4\textwidth]{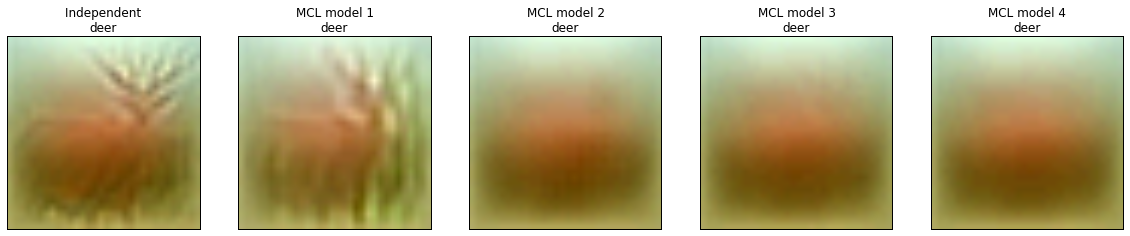}\hspace{20pt}
\includegraphics[width=0.4\textwidth]{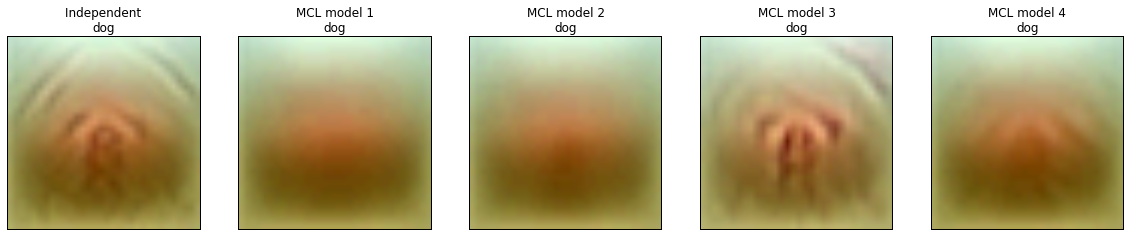}\\
\includegraphics[width=0.4\textwidth]{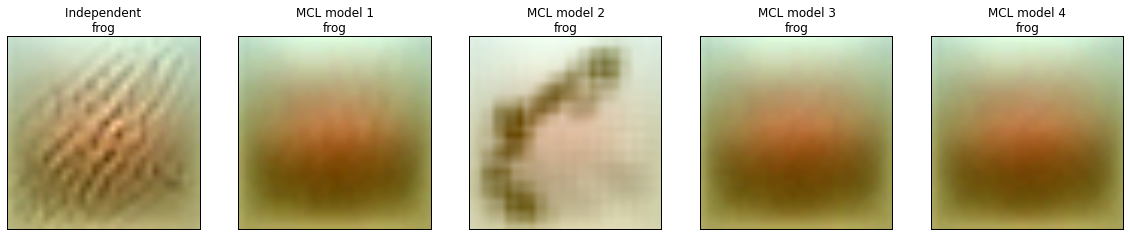}\hspace{20pt}
\includegraphics[width=0.4\textwidth]{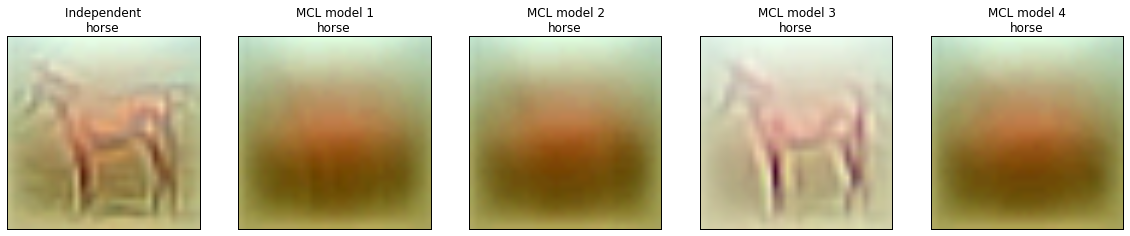}\\
\includegraphics[width=0.4\textwidth]{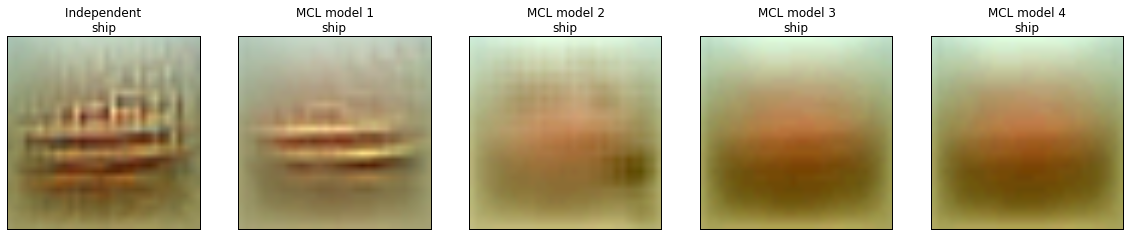}\hspace{20pt}
\includegraphics[width=0.4\textwidth]{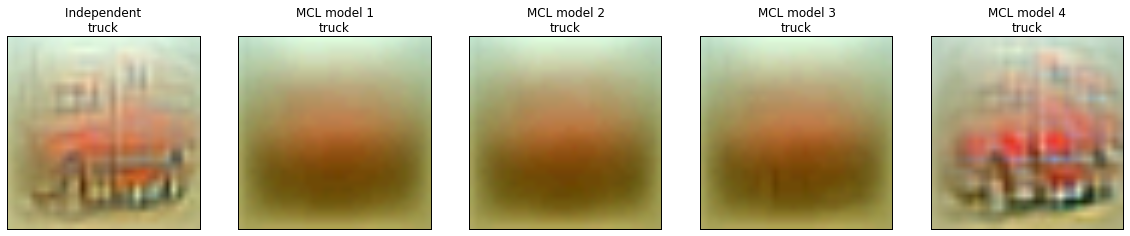}\\
\captionof{figure}{Reconstructions using features from the output layer using the images that give highest activation for the single model. Column1: Model from Standard Ensemble, Column2-4: Members of ensemble trained using MCL Loss }
\label{fig:ip2_vis}
\end{minipage}

\begin{minipage}{0.95\textwidth}
\vspace{30pt}
\center
\includegraphics[width=0.4\textwidth]{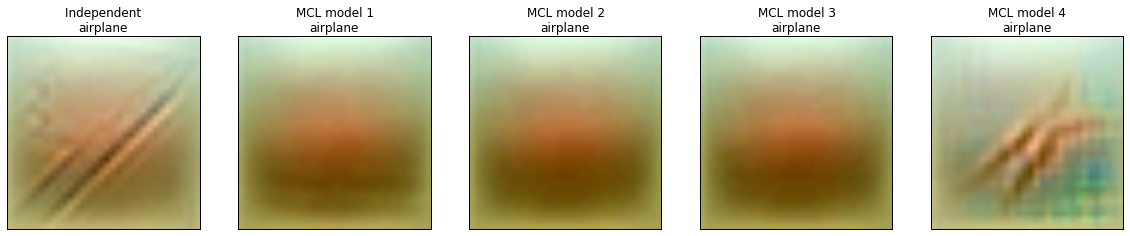}\hspace{20pt}
\includegraphics[width=0.4\textwidth]{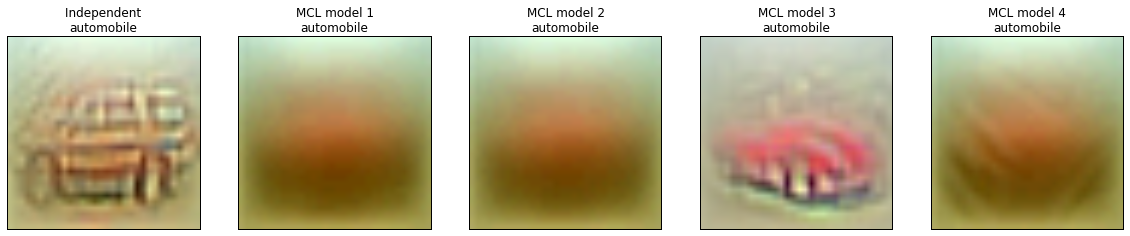}\\
\includegraphics[width=0.4\textwidth]{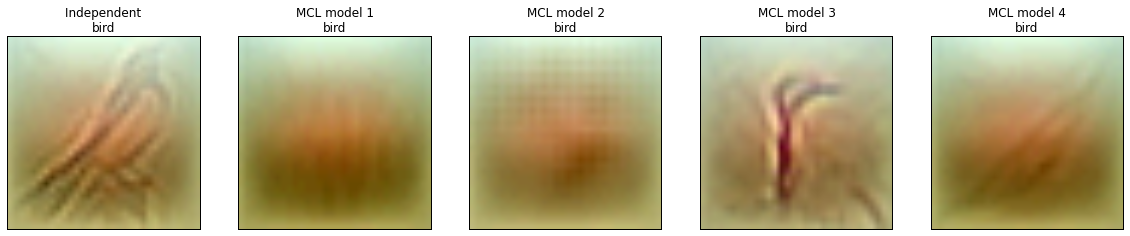}\hspace{20pt}
\includegraphics[width=0.4\textwidth]{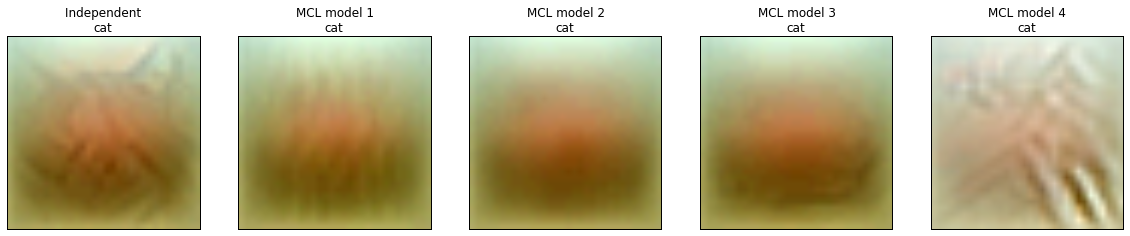}\\
\includegraphics[width=0.4\textwidth]{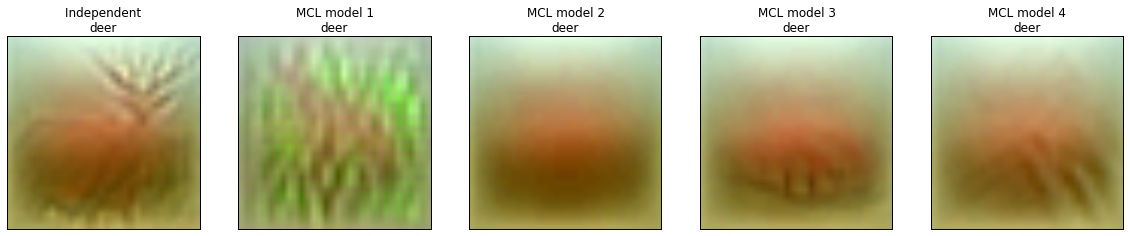}\hspace{20pt}
\includegraphics[width=0.4\textwidth]{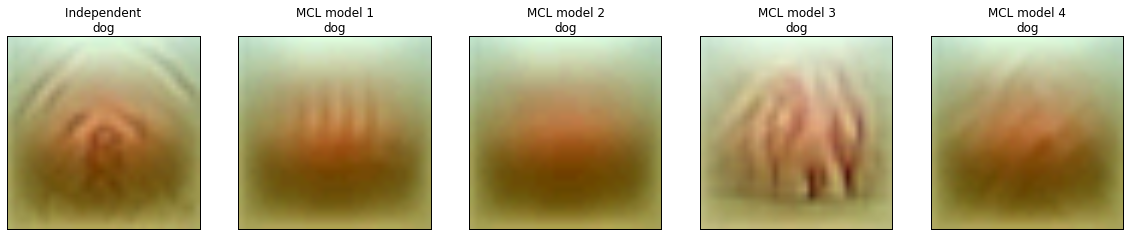}\\
\includegraphics[width=0.4\textwidth]{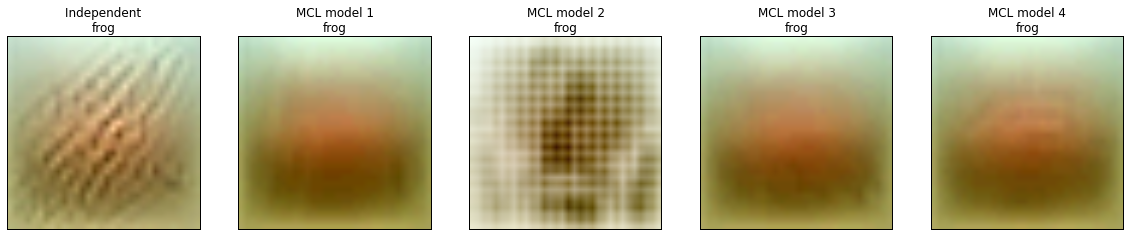}\hspace{20pt}
\includegraphics[width=0.4\textwidth]{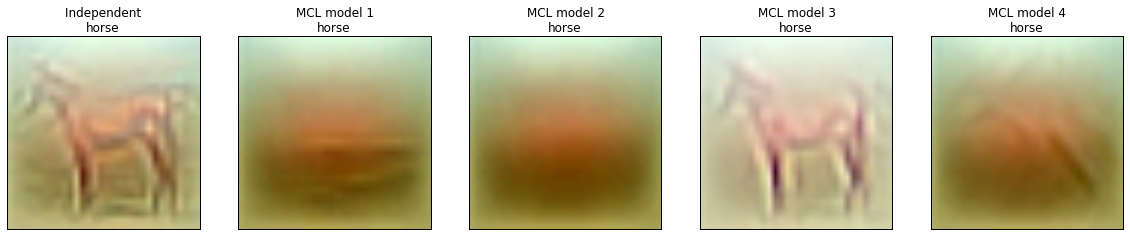}\\
\includegraphics[width=0.4\textwidth]{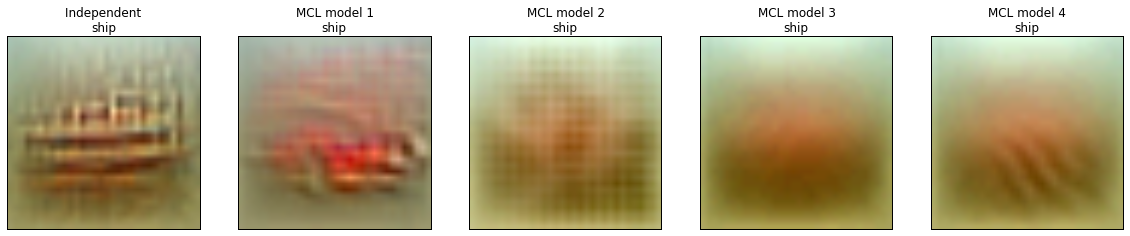}\hspace{20pt}
\includegraphics[width=0.4\textwidth]{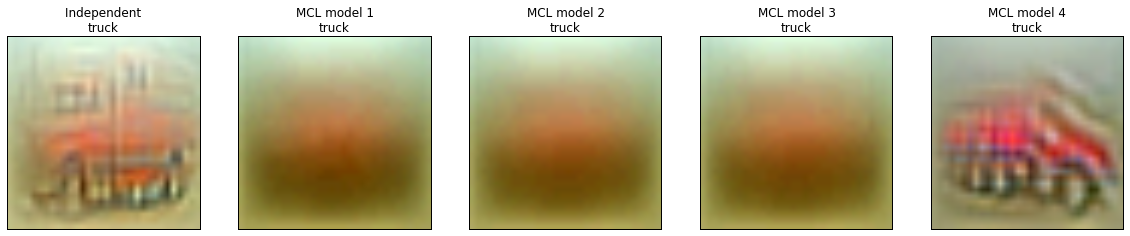}\\
\captionof{figure}{Reconstructions using features from the output layer using the images that give highest activation for each model independently. Column1: Model from Standard Ensemble, Column2-4: Members of ensemble trained using MCL Loss }
\label{fig:ip2_vis_indep}
\end{minipage}

\vspace{70pt}

\begin{minipage}{0.95\textwidth}
\center
\includegraphics[width=0.4\textwidth]{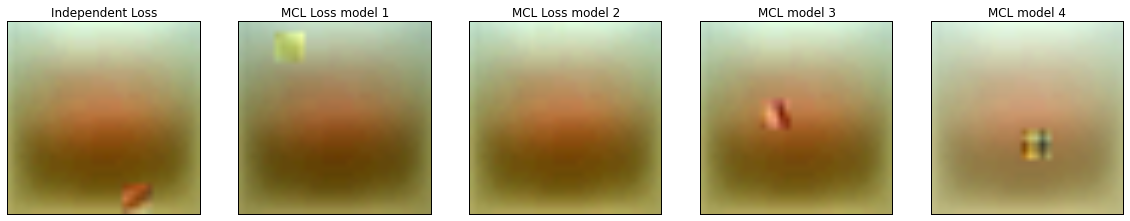}\hspace{20pt}
\includegraphics[width=0.4\textwidth]{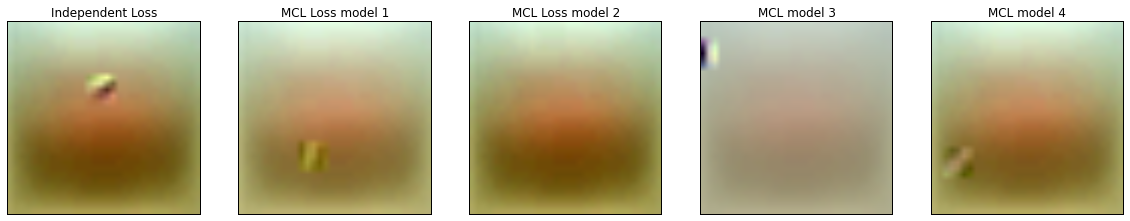}\\
\includegraphics[width=0.4\textwidth]{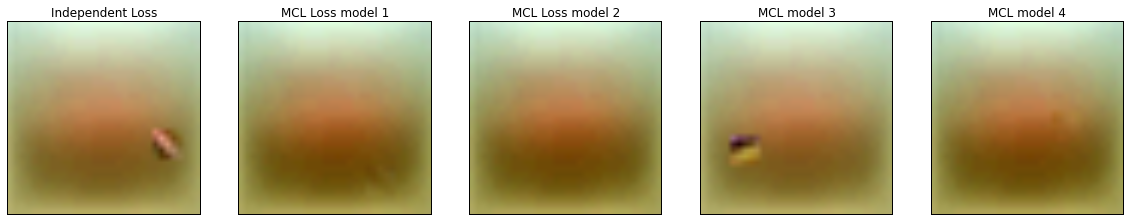}\hspace{20pt}
\includegraphics[width=0.4\textwidth]{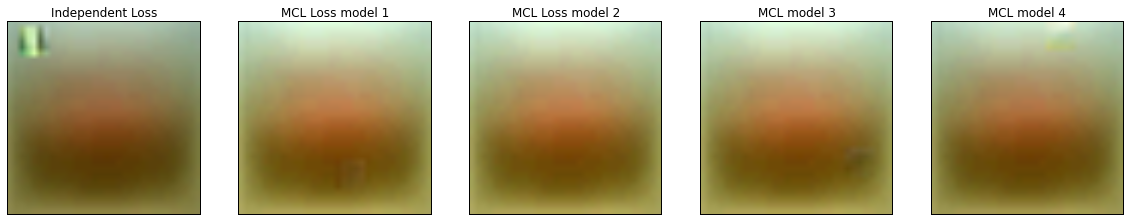}\\
\includegraphics[width=0.4\textwidth]{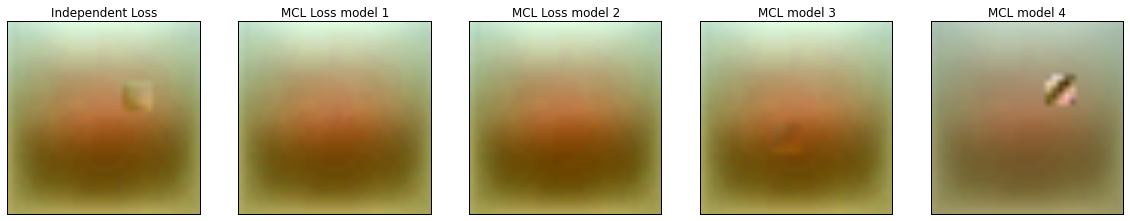}\hspace{20pt}
\includegraphics[width=0.4\textwidth]{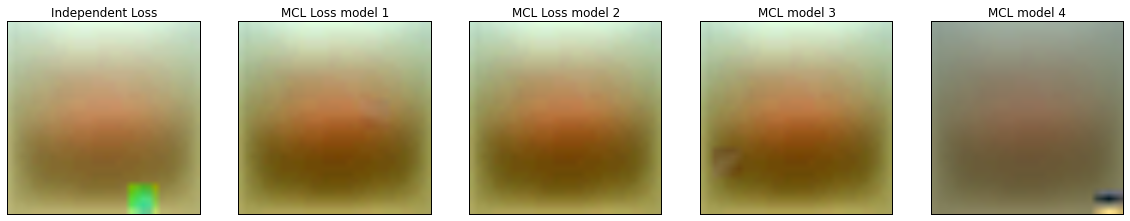}\\
\includegraphics[width=0.4\textwidth]{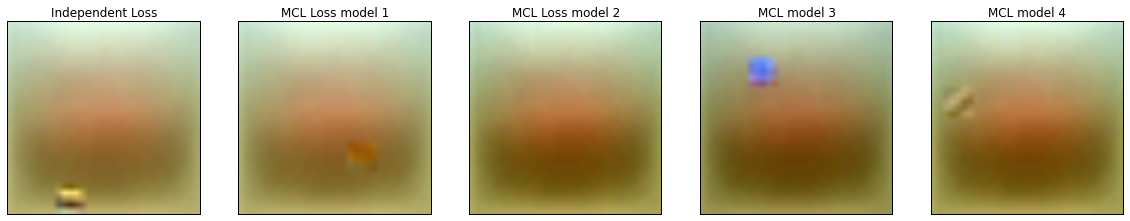}\\
\captionof{figure}{Reconstructions using the conv1 layer. Column1: Model from Standard Ensemble, Column2-4: Members of ensemble trained using MCL Loss }
\label{fig:conv1_vis}
\end{minipage}

\begin{minipage}{0.95\textwidth}
\vspace{30pt}
\center
\includegraphics[width=0.4\textwidth]{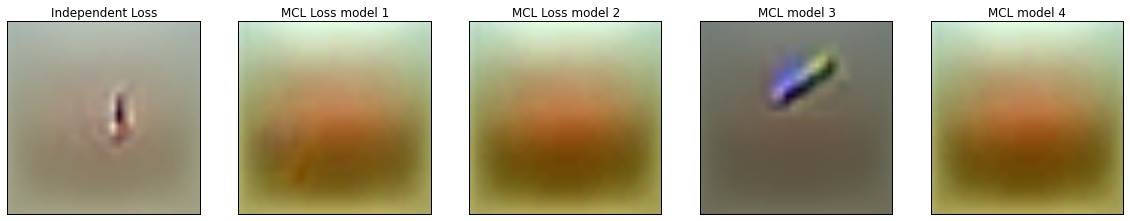}\hspace{20pt}
\includegraphics[width=0.4\textwidth]{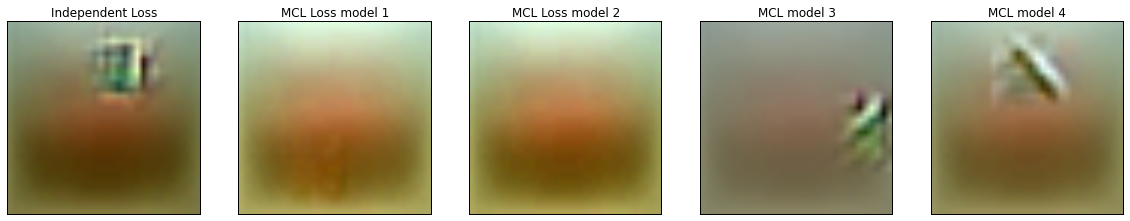}\\
\includegraphics[width=0.4\textwidth]{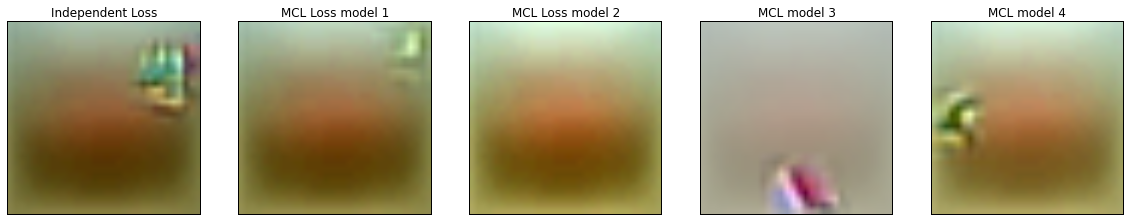}\hspace{20pt}
\includegraphics[width=0.4\textwidth]{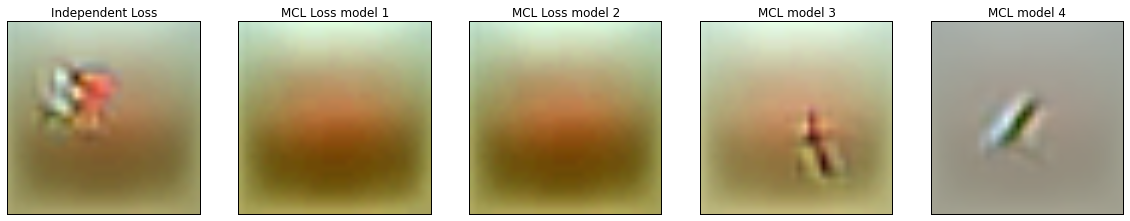}\\
\includegraphics[width=0.4\textwidth]{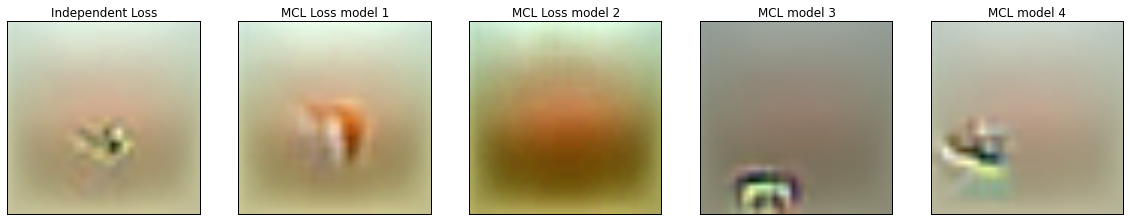}\hspace{20pt}
\includegraphics[width=0.4\textwidth]{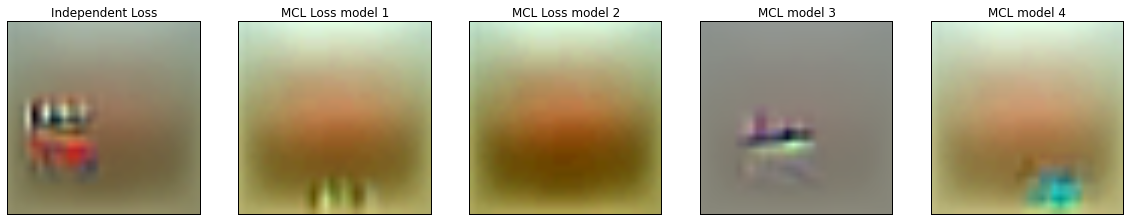}\\
\includegraphics[width=0.4\textwidth]{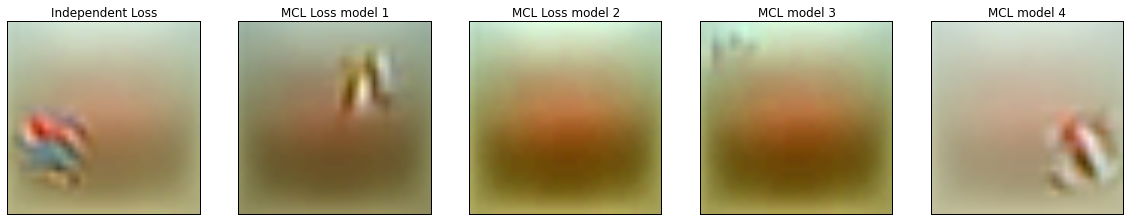}\hspace{20pt}
\includegraphics[width=0.4\textwidth]{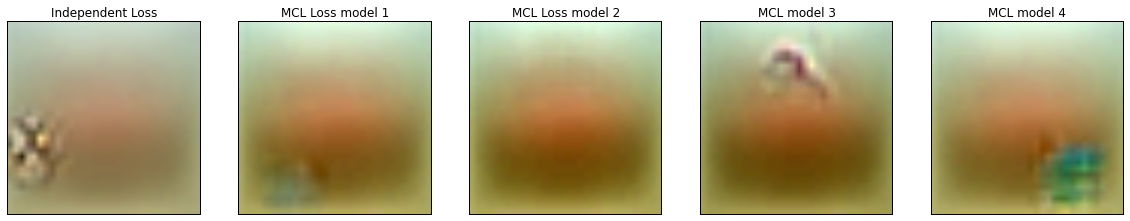}\\
\captionof{figure}{Reconstructions using the conv2 layer. Column1: Model from Standard Ensemble, Column2-4: Members of ensemble trained using MCL Loss }
\label{fig:conv2_vis}
\end{minipage}

\vspace{70pt}

\begin{minipage}{0.95\textwidth}
\center
\includegraphics[width=0.4\textwidth]{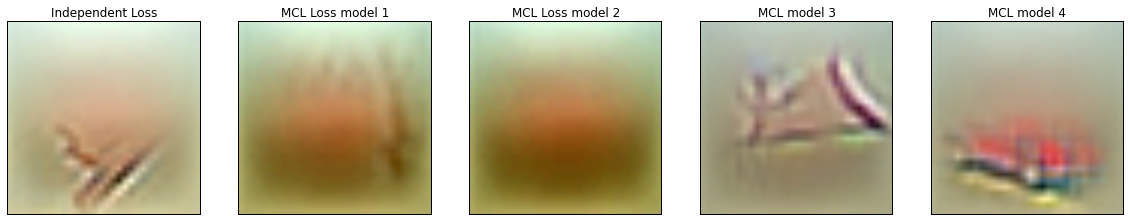}\hspace{20pt}
\includegraphics[width=0.4\textwidth]{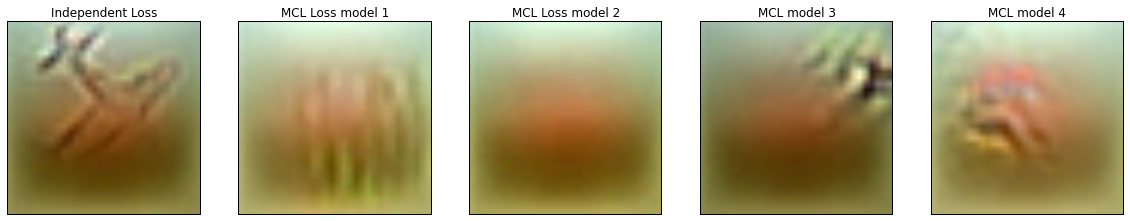}\\
\includegraphics[width=0.4\textwidth]{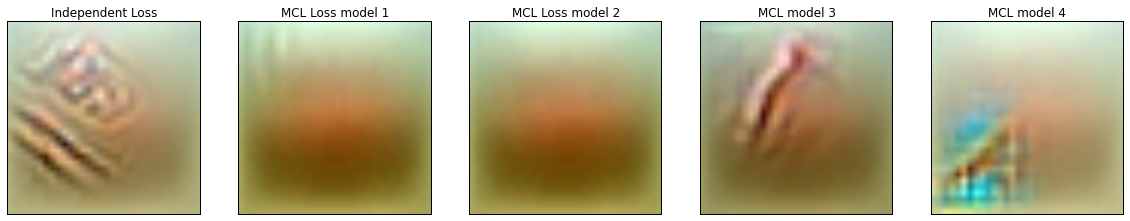}\hspace{20pt}
\includegraphics[width=0.4\textwidth]{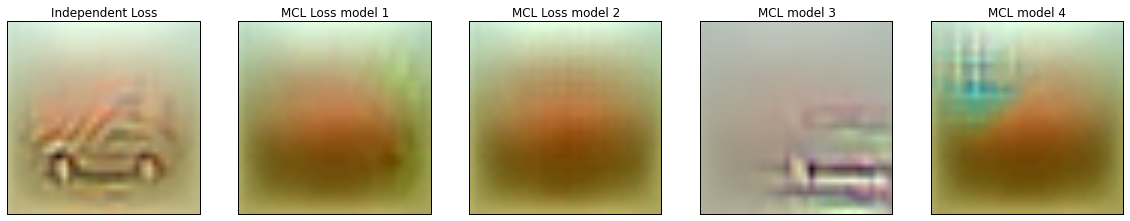}\\
\includegraphics[width=0.4\textwidth]{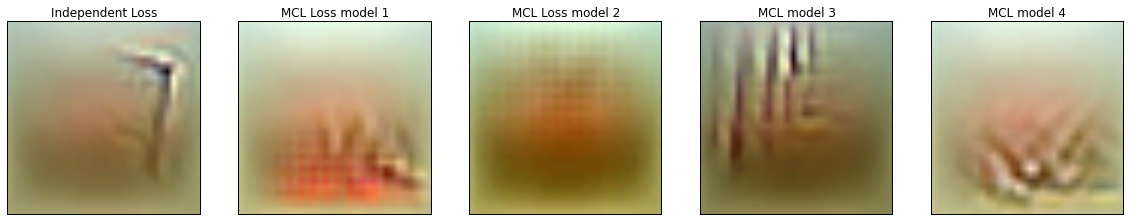}\hspace{20pt}
\includegraphics[width=0.4\textwidth]{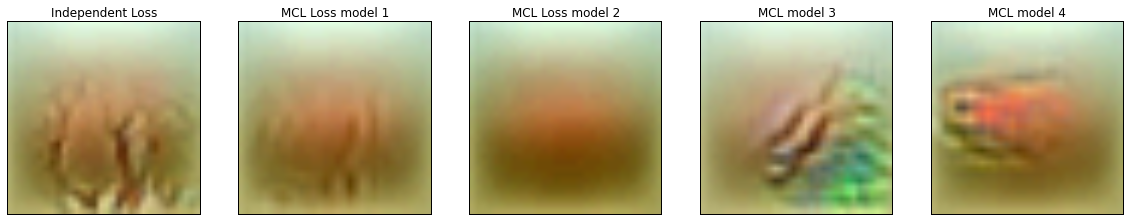}\\
\includegraphics[width=0.4\textwidth]{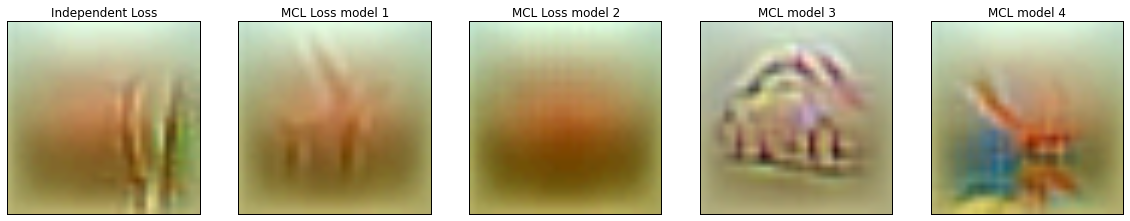}\hspace{20pt}
\includegraphics[width=0.4\textwidth]{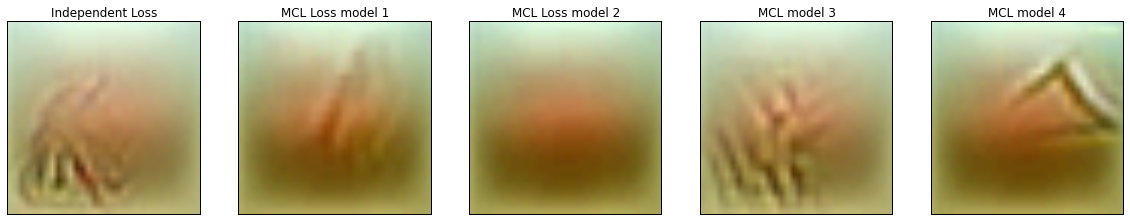}\\
\captionof{figure}{Reconstructions using the conv3 layer. Column1: Model from Standard Ensemble, Column2-4: Members of ensemble trained using MCL Loss }
\label{fig:conv3_vis}
\end{minipage}

\newpage
\section{MPI-Caffe}\label{app:mpi}
\noindent\texttt{MPI-Caffe} is a modification of the popular
\texttt{Caffe} deep learning framework that enables
cross-GPU/cross-machine communication on MPI enabled systems as model
layers. Providing these MPI operations as layers allows network
designers the flexibility to quickly experiment with distributed
networks while abstracting away much of the communication logic. This
enables experimentation with extremely large (i.e.~larger than can be
held in a single GPU) networks as well as ensemble-aware model
parallelism schemes. This document explains the function of these
layers as well as providing example usage. The core functionality in
\texttt{MPI-Caffe} is provided by
\begin{itemize}
\item the \texttt{MPIBroadcast} layer discussed in Section \ref{sec:broad} 
\item and the \texttt{MPIGather} layer discussed in Section \ref{sec:gather}.
\end{itemize}
The primary file defining the interface of the MPI layers is \texttt{MPILayers.hpp}. There are also many supporting modifications in the source that should be noted in case anyone tries to modify or update the base \texttt{Caffe} version. The network initialization code in \texttt{net.cpp} has been substantially altered to accommodate the distributed framework. Some other changes occur in \texttt{layer.hpp}, \texttt{solver.cpp}, and \texttt{caffe.cpp} among others.

\subsection{A Toy Example}
\noindent Let's start with a toy example to build context for the MPI layer descriptions. Suppose we want to train an TreeNet ensemble of CIFAR10-Quick and we want train it under a score-averaged loss.  Figure \ref{fig:lenet} shows how we might modify the LeNet structure using \texttt{MPI-Caffe} to implement this model across three processes/GPUs in an MPI enabled cluster. We will go through this example to explain the function and parametrization of the \texttt{MPIBroadcast} and \texttt{MPIGather} layers.

\begin{minipage}{\textwidth}
\centering
\includegraphics[trim=0px 0 15px 0, clip=true, width=0.7\columnwidth]{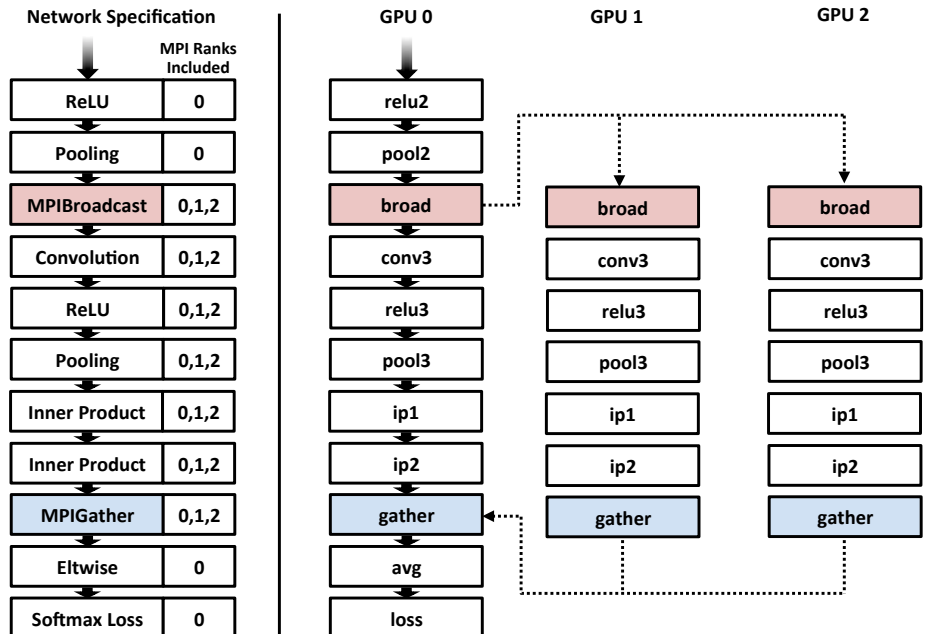}
\captionof{figure}{\textbf{Left)} Example network specification for \texttt{MPI-Caffe} enabled CIFAR10-Quick ensemble with parameter sharing and model averaging. \textbf{Right)} Distributed model resulting from the specification, visualized for each participating process. Dashed lines indicate MPI cross process communication and not layer input/outputs.}
\label{fig:lenet}
\end{minipage}

\subsubsection{MPIBroadcast}
\label{sec:broad}
\noindent The first layer we discuss is 
\texttt{MPIBroadcast}  (highlighted in red in Figure
\ref{fig:lenet}). The \texttt{MPIBroadcast} layer broadcasts a copy of
its input blob to each process in its communication group during its
forward pass. During the backward pass, the gradients from each copy
are summed and passed back to the input blob. The communication group
consists of all processes that carry a copy of a particular broadcast
layer. By default a communication group contains all processes;
however, adding \texttt{mpi\_rank:n} rules in either the
\texttt{include} or \texttt{exclude} layer parameters can alter this
group.

\begin{wrapfigure}{l}{0.5\textwidth}
\begin{minipage}{0.5\textwidth}
\vspace{10pt}
\centering
\BUseVerbatim[fontshape=tt,fontsize=\scriptsize,fontfamily=courier,commandchars=\\\{\}]{BROAD}
\includegraphics[scale=0.29]{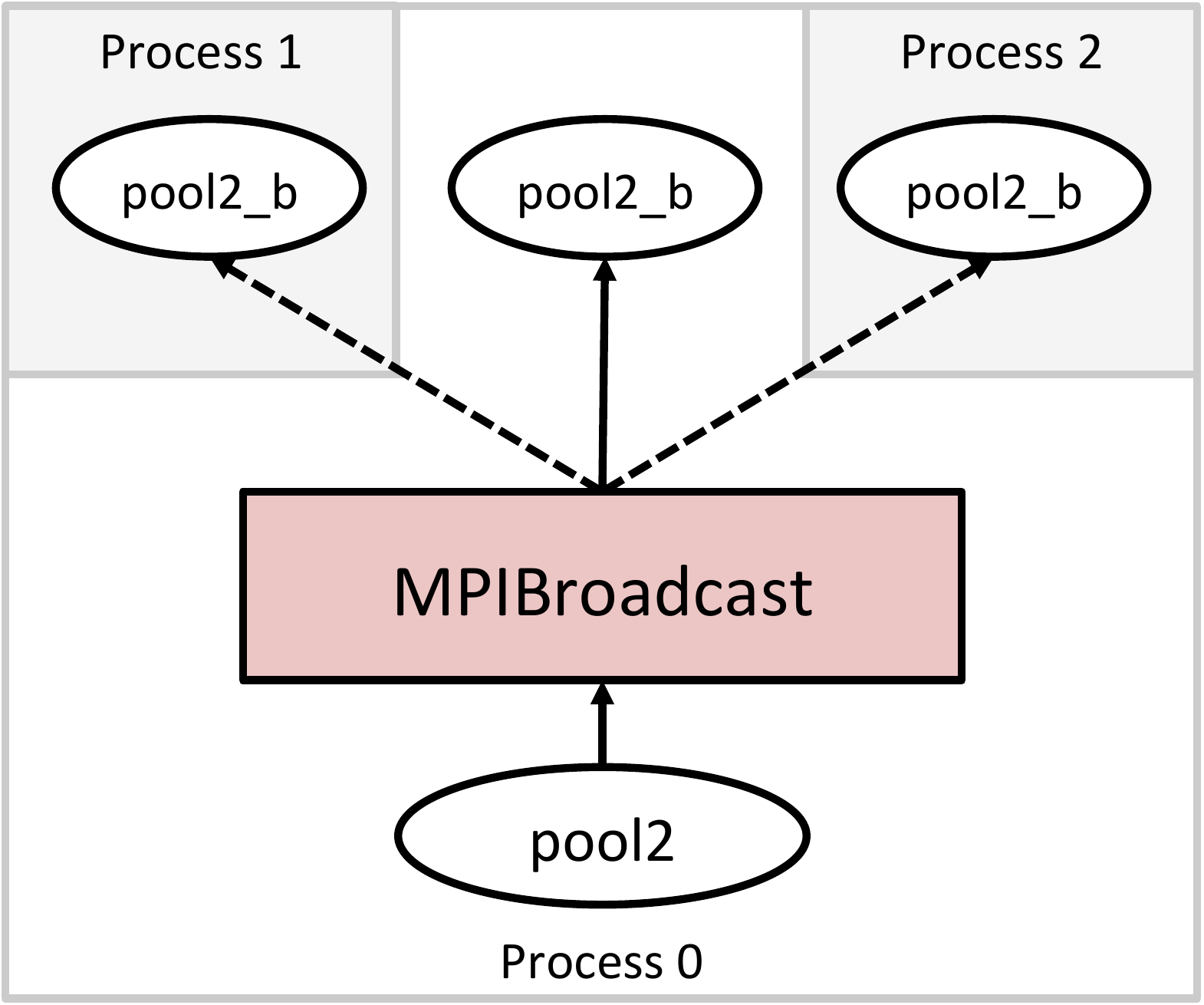}
\captionof{figure}{An example \texttt{MPIBroadcast} layer definition and a diagram of its forward-pass behavior.}
\label{fig:broad}
\end{minipage}
\vspace{-20pt}
\end{wrapfigure}

Figure \ref{fig:broad} shows the \texttt{MPIBroadcast} layer definition from our example and the corresponding forward-pass behavior. Going step by step through the definition: we
\begin{itemize}
\item \textcolor{red}{\texttt{[Line 2-5]} declare this layer to be a \texttt{MPI} \texttt{Broadcast} layer named ``broad'' with input blob \texttt{pool2} and output blob \texttt{pool2\_b},}
\item \textcolor{darkorchid}{\texttt{[Line 6-8]} set the \texttt{mpi\_param} root value to 0 indicating that process 0 will be initiating the broadcast,}
\item  \textcolor{blue}{and \texttt{[Line 9-13]} establish a communication group consisting of processes 0, 1, and 2.}
\end{itemize}
During a forward pass, the \texttt{MPIBroadcast} layer on process 0 will send a copy of \texttt{pool2} to processes 1 and 2 as well as retain a copy for itself. For the example, we would also need to modify the \texttt{ip1} layer to take \texttt{pool2\_b} as input rather than \texttt{pool2}.

It is important to note the effect the choice of \texttt{mpi\_param\{ root \}} has on network structure. As shown in the example in Figure \ref{fig:lenet}, each process parses the entire network structure and retains only the layers that include its MPI rank. For process 0, this includes the entire network, but for processes 1 and 2 the network \emph{starts} with the \texttt{MPIBroadcast} layer. In order to allow this behavior, non-root processes have the input blob (\texttt{pool2} in our example) stripped out during network parsing. Additionally for this example we need to average these top blobs before sending the result into the softmax loss.

\subsubsection{MPIGather}
\label{sec:gather}
\begin{wrapfigure}{r}{0.5\textwidth}
\vspace{-10pt}
\begin{minipage}{0.5\textwidth}
\centering
\begin{subfigure}[t]{0.4\columnwidth}
\centering
\BUseVerbatim[fontshape=tt,fontsize=\scriptsize,fontfamily=courier]{GATHER}
\end{subfigure}%
\begin{subfigure}[b]{0.6\columnwidth}
\centering
\includegraphics[scale=0.29]{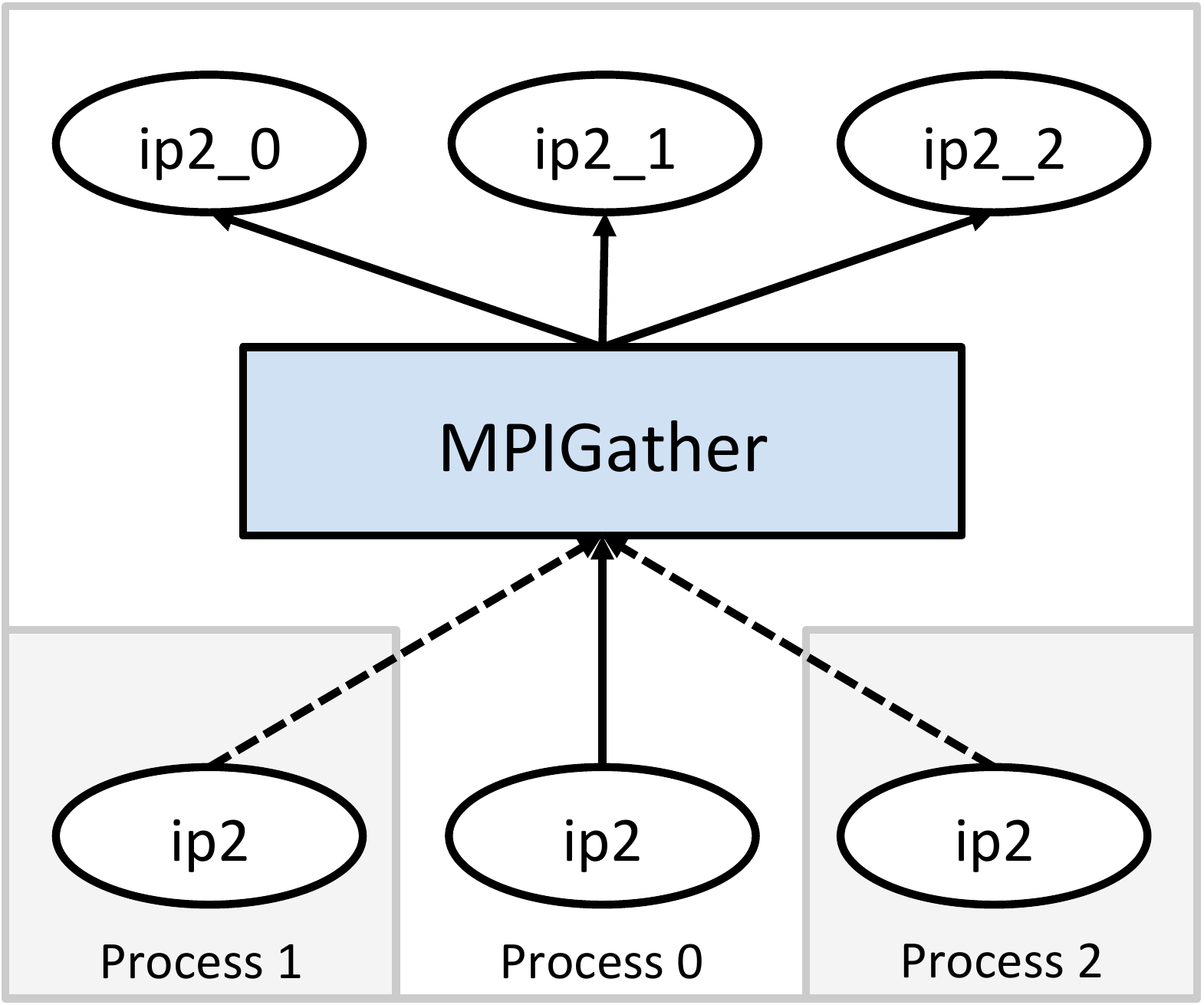}
\vspace{13pt}
\end{subfigure}
\caption{An example \texttt{MPIGather} layer definition and a diagram of its forward-pass behavior.}
\label{fig:gather}
\end{minipage}
\end{wrapfigure}

\noindent If the purpose of a broadcast layer is to take some data and push copies into multiple process spaces, the \texttt{MPIGather} layer can be thought of as the opposite. In a forward pass, it takes multiple copies of a blob from multiple process spaces and collects them in the root process. During a backward pass, the gradients for each top blob are routed back to the corresponding input blob and process. Similar to the previous section, Figure \ref{fig:gather} shows the layer definition from out example and a diagram of the forward pass behavior.

\begin{SaveVerbatim}[]{GATHER}
1 layer{
2  name: ExampleLayer
3  type: MPIGather
4  bottom: ip2
5  top: ip2_0
6  top: ip2_1
7  top: ip2_2
8  mpi_param{
9    root: 0
10  }
11 include{
12    mpi_rank: 0
13    mpi_rank: 1
14    mpi_rank: 2
15  }
16 }
\end{SaveVerbatim}

The \texttt{mpi\_param\{root\}} parameter in the gather layer defines
which process will be receiving the gathered blobs and producing top
blobs. In analogy to the broadcast layer parsing, gather layers in
non-root processes are pruned of the top blobs during network parsing
(see Figure \ref{fig:lenet}).

There are some restrictions to the gather layer's use. First, the
bottom blob (\texttt{ip2} in our example) must be defined in all
communication group processes. Second, the number of top blobs must
equal the number of processes in the communication group. Both of
these conditions are checked by the source and will report an error if
not satisfied.

\subsection{Notes and Other Examples}
\noindent It is worth noting a few other use points about \texttt{MPI-Caffe}:
\begin{itemize}
\item the \texttt{MPIBroadcast} layer can be used to construct a very large single-path network spanned across multiple GPU's 
\item the \texttt{MPIGather} layer can be used to allow more sophisticated ensemble losses
\item there is no limit on the number or order of MPI layers such that complex distributed networks are possible
\item in situations where network latency is lower than reading from disk, the \texttt{MPIBroadcast} layer can be used to train multiple independent networks more quickly
\end{itemize}

\subsection{Communication Cost Analysis}
\noindent We tested our MPI-Caffe framework on a large-scale cluster with one Tesla K20
GPU per node and a maximum MPI node interconnect bandwidth of 5.8 GB/sec. 
To characterize the communication overhead for an
ensemble, we measure the time spent sharing various layers of the
ILSVRC-Alex{\footnotesize $\times$5} architecture. Each network was run on a separate node (with one node also holding the shared layers). Figure \ref{fig:mpi} shows the communication time to share a given layer as a fraction of the forward-backward pass. The x-axis indicates the number of floats broadcast per batch for each layer. We note that for these layers overhead appears approximately linear and even the largest layer incurs very little overhead for communication.  

\begin{minipage}{0.95\textwidth}
\vspace{12pt}
\centering
\includegraphics[clip=true, trim=5px 5px 5px 5px, width=0.75\columnwidth]{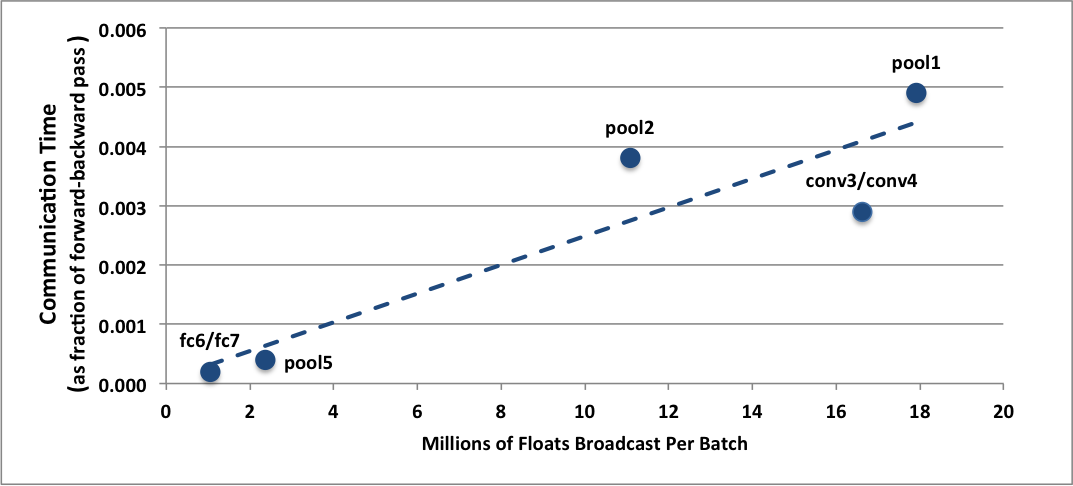}
\captionof{figure}{Fraction of forward-backward pass time used for TreeNets sharing various layers against the size of the layers. The overhead from communication is quite small and scales approximately linearly with the size of the layer being shared.}
\label{fig:mpi}
\end{minipage}

\end{document}